\documentclass[pdflatex,sn-vancouver-ay,iicol]{sn-jnl}% Math and Physical Sciences Numbered Reference Style
\usepackage{graphicx}%
\usepackage{multirow}%
\usepackage{amsmath,amssymb,amsfonts}%
\usepackage{amsthm}%
\usepackage{mathrsfs}%
\usepackage[title]{appendix}%
\usepackage{xcolor}%
\usepackage{textcomp}%
\usepackage{manyfoot}%
\usepackage{booktabs}%
\usepackage{algorithm}%
\usepackage{algorithmicx}%
\usepackage{algpseudocode}%
\usepackage{listings}%
\usepackage{bm}
\usepackage{mathtools}
\usepackage{enumitem}
\usepackage{subcaption}

\usepackage{stfloats}
\usepackage{tabularx}
\usepackage{booktabs}

\usepackage{multirow}
\usepackage{booktabs}
\usepackage{adjustbox}
\usepackage{lscape}

\usepackage{booktabs}
\usepackage{graphicx}

\newtheorem{theorem}{Theorem}[section]

\newtheorem{corollary}[theorem]{Corollary}
\newtheorem{proposition}[theorem]{Proposition}
\theoremstyle{definition}
\newtheorem{definition}[theorem]{Definition}
\newtheorem{remark}[theorem]{Remark}

\newcommand{\R}{\mathbb{R}}
\newcommand{\E}{\mathbb{E}}
\newcommand{\Var}{\operatorname{Var}}
\newcommand{\tr}{\operatorname{tr}}
\newcommand{\norm}[1]{\lVert#1\rVert}
\newcommand{\abs}[1]{\lvert#1\rvert}
\newcommand{\eps}{\varepsilon}
\newcommand{\rvec}{\bm{r}}
\newcommand{\uvec}{\bm{u}}

\newcommand{\SNR}{\mathrm{SNR}}

\begin{document}
	
	\title[MS-DGCNN++: Multi-Scale Dynamic Graph Convolution with Scale-Dependent Normalization for Robust LiDAR Tree Species Classification]{MS-DGCNN++: Multi-Scale Dynamic Graph Convolution with Scale-Dependent Normalization for Robust LiDAR Tree Species Classification}
	
	%%=============================================================%%
	%% GivenName	-> \fnm{Joergen W.}
	%% Particle	-> \spfx{van der} -> surname prefix
	%% FamilyName	-> \sur{Ploeg}
	%% Suffix	-> \sfx{IV}
	%% \author*[1,2]{\fnm{Joergen W.} \spfx{van der} \sur{Ploeg} 
		%%  \sfx{IV}}\email{iauthor@gmail.com}
	%%=============================================================%%
	
	\author*[1]{\fnm{Said} \sur{Ohamouddou}}\email{said\_ohamouddou@um5.ac.ma}
	
	\author[1]{\fnm{Hanaa} \sur{El Afia}}
	
	\author[1]{\fnm{Mohamed Hamza} \sur{Boulaich}}
	
	\author[1]{\fnm{Abdellatif} \sur{El Afia}}
	
	\author[1]{\fnm{Raddouane} \sur{Chiheb}}
	
	\affil*[1]{\orgdiv{ENSIAS}, 
		\orgname{Mohammed V University}, 
		\orgaddress{
			\street{Avenue Mohammed Ben Abdallah Regragui, Madinat Al Irfane, BP 713, Agdal},
			\city{Rabat},
			\country{Morocco}
	}}
	
	%%==================================%%
	%% Sample for unstructured abstract %%
	%%==================================%%
	
	\abstract{
		Graph-based deep learning on LiDAR point clouds encodes local geometry through edge features from neighboring displacements; however, standard implementations use the same encoding at every scale.
		In tree species classification, where point density varies by orders of magnitude between the trunk and canopy, this uniform treatment is particularly limiting.
		We show that it is suboptimal: normalized directional features have mean squared error decaying as $\mathcal{O}(1/s^2)$ with inter-point distance~$s$, while raw displacement error remains constant, implying that each encoding suits a different signal-to-noise ratio (SNR) regime. We propose MS-DGCNN++, a multi-scale dynamic graph convolutional network with \emph{scale-dependent edge encoding}: raw vectors at the local scale (small~$k$, low SNR) and hybrid raw-plus-normalized vectors at the intermediate scale (large~$k$, high SNR).
		Five ablations validate this design: encoding ablation confirms $+4$--$6\%$ overall accuracy (OA) gain; density dropout shows the flattest degradation under canopy thinning; a noise sweep locates the theoretical crossover near $\text{SNR}_2 \approx 1.22$; max-pooling provenance reveals far neighbors win $85\%$ of competitions under raw encoding, a bias eliminated by normalization; and isotropy analysis shows normalization nearly doubles effective rank. On STPCTLS (seven species, terrestrial laser scanning (TLS)), MS-DGCNN++ achieves the highest OA ($92.91\%$) among 56 models, surpassing self-supervised methods with $7$--$24\times$ more parameters using only $1.81$M parameters. On HeliALS (nine species, airborne laser scanning (ALS), geometry-only), it achieves $73.66\%$ OA with the best balanced accuracy ($50.28\%$), matching FGI-PointTransformer, which uses $4\times$ more points. Robustness analysis across five perturbation types revealed complementary variant strengths, providing deployment guidance for heterogeneous forest environments. The code used in this study is available at \url{https://github.com/said-ohamouddou/MS-DGCNN2}.
	}

	%%\pacs[JEL Classification]{D8, H51}
	
	%%\pacs[MSC Classification]{35A01, 65L10, 65L12, 65L20, 65L70}
	
	\maketitle

	\section{Introduction}
	\label{sec:intro}
	
	Three-dimensional point clouds have become a central data modality in computer vision, robotics, and remote sensing.
	Their irregular, unordered, and density-varying nature poses fundamental challenges for representation learning, and a rich landscape of methods has emerged to address these challenges.
	\citep{qi2017pointnet} established the first deep architecture that directly processes raw point clouds while respecting permutation invariance, and \citep{qi2017pointnet++} extended this with hierarchical feature learning at multiple spatial scales.
	Subsequent point-based methods have enriched local feature extraction through learned point transformations~\citep{pointcnn}, relational shape descriptors~\citep{rscnn}, and improved training strategies~\citep{qian2022pointnext}.
	Self-attention mechanisms have also been adapted to this domain: the Point Transformer~\citep{zhao2021point} and PCT~\citep{guo2021pct} introduced attention for point-wise feature learning, whereas more recent variants have achieved state-of-the-art results through serialization-based attention~\citep{wu2024point} and state-space models~\citep{liang2024pointmamba}.
	Self-supervised pre-trainingvia masked reconstruction~\citep{pang2022masked,zhang2022point,qi2023contrast,chen2024pointgpt} or contrastive learning~\citep{xie2020pointcontrast,afham2022crosspoint}has further improved data efficiency, although at the cost of large transformer backbones ($12$--$44$M parameters) and pretraining on external datasets such as ShapeNet~\citep{chang2015shapenet}.
	Parameter-efficient fine-tuning strategies, such as prompt tuning~\citep{sun24ppt}, adapter modules~\citep{zhou2024dynamic}, and graph-based self-training~\citep{PointGST}, reduce the number of \emph{trainable} parameters during adaptation. However, the full pre-trained backbone must still be loaded and executed during inference, leaving the deployment footprint unchanged.
	
	Despite the success of these large-scale approaches, lightweight architectures remain essential for applications in which inference latency, memory footprint, and energy consumption are binding constraints.
	A complementary line of research models point clouds as graphs, explicitly encoding the topological and geometric relationships among points.
	DGCNN~\citep{wang2019dynamic} introduced EdgeConv, which dynamically recomputes $k$-nearest-neighbor graphs in the feature space at each layer, capturing both local geometry and non-local semantic similarities as features evolve.
	Graph-based approaches offer a compelling inductive bias: they are naturally permutation-invariant, operate in $\mathcal{O}(nk)$ rather than the $\mathcal{O}(n^2)$ of full self-attention, and require far fewer parameters, making them particularly suitable for deployment on resource-constrained edge devices.
	
	\vspace{1mm}
	\noindent\textbf{The multi-scale challenge.}
	However, a central limitation of DGCNN is its reliance on a single, fixed neighborhood size~$k$ at each layer.
	While adequate for synthetic benchmarks such as ModelNet40, where objects are uniformly sampled, this becomes a significant bottleneck when point clouds exhibit strong \emph{geometric heterogeneity}---when different regions of the same object contain structures at fundamentally different scales.
	This challenge is particularly acute in forestry applications using terrestrial laser scanning (TLS) and airborne laser scanning (ALS) data, where tree point clouds present an extreme multi-scale structure: dense trunk surfaces ($k{=}5$--$10$ sufficient) to sparse, stochastic canopy geometry ($k{=}20$--$40$ needed)~\citep{liuMultifeatureFusionNetwork2025}.
	A single~$k$ cannot simultaneously resolve fine-grained surface textures or capture broad shape silhouettes. This constitutes an architectural limitation that cannot be resolved by hyperparameter selection alone.
	
	This limitation manifests itself in the empirical performance.
	On ModelNet40, a synthetic benchmark with uniformly sampled CAD models, DGCNN achieves $92.9\%$ OA, outperforming PointNet++ MSG ($91.9\%$)~\citep{wang2019dynamic}.
	However, the situation is reversed for geometrically heterogeneous real-world data.
	On FOR-species20K, ensemble multi-scale PointNet++ reached $75.6\%$ OA versus $68.3\%$ for DGCNN~\citep{pulitiBenchmarkingTreeSpecies2025a}; on mobile laser scanning data, PointNet++ achieved an F1 score of $97.31\%$ compared to $96.47\%$ for DGCNN~\citep{liu2022treespecies_mls}; and on STPCTLS, MFFTC-Net---a multi-feature fusion method---outperformed DGCNN at $90.37\%$ versus $87.41\%$ as reported by~\citep{liuMultifeatureFusionNetwork2025} under their training protocol.
	The common thread is that methods incorporating multi-scale or multi-feature representations consistently surpass the DGCNN on tree species data, where non-uniform density, noise, missing points, and outliers are inherent.
	This reversal isolates the single-scale bottleneck as the limiting factor and motivates the multi-scale extension of the DGCNN.
	
	\vspace{1mm}
	\noindent\textbf{Why existing multi-scale extensions are insufficient.}
	The multi-scale dynamic DGCNN (MS-DGCNN) of \citep{zhaiMultiScaleDynamicGraph2020c} addressed this by running three parallel EdgeConv paths at $k{=}20, 30, 40$.
	While this improved results on ModelNet10, gains on more challenging datasets were marginal---the authors' own published results show that on ModelNet40, the single-scale version outperformed the proposed multi-scale variant.
	We identified four design deficiencies.
	\emph{First, redundant edge encoding:} all three paths use the same raw edge features, producing overlapping rather than complementary representations.
	\emph{Second, no cross-scale interaction:} the parallel paths process independently, unable to model hierarchical dependencies between fine-grained local detail and broader structural patterns.
	\emph{Third, no theoretical foundation:} the study provides no formal justification for why multi-scale processing should help, no ablation of the number of scales, and no analysis of how the choice of $k$ values affects performance.
	\emph{Fourth, density sensitivity:} at larger neighbourhood radii, raw edge magnitudes are dominated by point spacing variation rather than intrinsic geometry---a confound that worsens in real-world acquisitions where density can vary by orders of magnitude.
	
	\vspace{1mm}
	\noindent\textbf{Our approach: MS-DGCNN++.}
	We propose MS-DGCNN++, a hierarchical multi-scale dynamic graph convolutional network that addresses all four deficiencies through two principled design choices.
	
	\textbf{Scale-dependent edge encoding.}
	Instead of applying the same edge features at all scales, MS-DGCNN++ uses a \emph{hybrid} encoding at the intermediate scale ($k_2$): the raw relative vector $\bm{r} = q - p$ is concatenated with the normalized direction $\hat{\bm{r}} = \bm{r}/\|\bm{r}\|$ and absolute position~$p$, producing a 9-channel input.
	At the local scale ($k_1 < k_2$), only the raw relative vector and position are used (six channels).
	This asymmetric design is grounded in a noise-sensitivity analysis: normalized directional features have Mean Squared Error (MSE) decaying as $\mathcal{O}(1/s^2)$ with inter-point distance~$s$, whereas raw features have constant MSE.
	Therefore, the hybrid encoding provides density-invariant angular descriptors, where density variation is the dominant confound (intermediate scale), while preserving fine-grained metric geometry, where directional estimates are noisy (local scale).
	
	\textbf{Hierarchical feature fusion.}
	The local and intermediate-scale features are fused through a learned convolution before being passed to the subsequent EdgeConv layers operating in the learned feature space at a global scale ($k_3$).
	This creates a progressive refinement pipeline---local surface detail $\to$ intermediate-scale angular structure $\to$ global semantic features---in contrast to the flat, parallel processing of MS-DGCNN.
	
	\vspace{1mm}
	\noindent This paper makes the following key contributions:
	\begin{enumerate}
		\item \textbf{Architecture.} We propose the MS-DGCNN++, a hierarchical multi-scale dynamic graph convolutional network with scale-dependent edge encoding.
		The design introduces normalized directional features at the intermediate scale and raw metric features at the local scale and is applicable to any point-cloud domain exhibiting geometric heterogeneity.
		
		\item \textbf{Theoretical justification.} We provide a noise-sensitivity framework proving that hybrid encoding is noise-optimal: normalized directions have MSE decaying as $\mathcal{O}(1/s^2)$, raw vectors have constant MSE, and their concatenation exploits complementary noise profiles.
		Five targeted experiments---per-scale ablation, density dropout, noise sweep, max-pooling provenance analysis, and feature-space isotropy analysis---validated this framework.
		
		\item \textbf{Empirical results.} On STPCTLS~\citep{seidelPredictingTreeSpecies2021c} (seven species, TLS), MS-DGCNN++ achieved the highest overall accuracy ($92.91\%$) among 56 evaluated models, including self-supervised methods with $7$$24\times$ more parameters, using only $1.81$M parameters and no pre-training.
		On HeliALS~\citep{taher2026multispectral} (nine species, ALS), it achieved $73.66\%$ OA with the best balanced accuracy among all geometry-only methods, matching the FGI-PointTransformer baseline that uses $4\times$ more input points.
		A comprehensive robustness analysis across noise, density dropouts, outlier injections, point reductions, and few-shot regimes provides practical guidance for deployment.
	\end{enumerate}

	\subsection{Paper Organisation}
	
	Section~\ref{sec:background} reviews LiDAR-based tree species classification and DGCNN formalism.
	Section~\ref{sec:method} presents the MS-DGCNN++ architecture.
	Section~\ref{sec:robustness} presents a theoretical noise-sensitivity framework for scale-dependent edge encoding.
	Section~\ref{sec:exp} describes the experimental setup of this study.
	Section~\ref{sec:ablation} reports the ablation studies and robustness evaluations.
	Section~\ref{sec:sota} presents a comparison with state-of-the-art methods.
	Section~\ref{sec:discussion} discusses the results and their limitations.
	Section~\ref{sec:conclusion} concludes the paper.

	%=============================================================================
	\section{Related Work}
	\label{sec:background}
	%=============================================================================
	
	We focus on two bodies of work that directly underpin this study: LiDAR-based tree species classification (\S\ref{subsec:lidar_tree}), which exemplifies the geometric heterogeneity challenge, and multi-scale point cloud methods (\S\ref{subsec:multiscale_methods}), whose shared limitation motivates our scale-dependent encoding.
	The DGCNN formalism upon which MS-DGCNN++ is built is reviewed in \S\ref{subsec:dgcnn}.
	
	%---------------------------------------------------------------
	\subsection{LiDAR-Based Tree Species Classification}
	\label{subsec:lidar_tree}
	%---------------------------------------------------------------
	
	Tree species classification is essential for forest ecosystem management~\citep{paillet2010biodiversity} and serves as a demanding testbed for point-cloud understanding. Individual trees span centimeter-level bark texture to meter-level crown morphology, and the point density varies by several orders of magnitude between the trunk and canopy.
	Research in this area has evolved over three generations.
	
	\smallskip
	\noindent\textbf{Handcrafted and 2D projection methods.}
	Early work relied on engineered statistics: \citep{TERRYN2020170} extracted 17 structural features from quantitative structure models fitted to TLS data, achieving ${\approx}80\%$ accuracy on five deciduous species, and \citep{michalowskaReviewTreeSpecies2021a} showed that full-waveform-derived features yielded the highest accuracy on airborne LiDAR.
	Second-generation methods project point clouds into 2D images to leverage mature CNN backbones.
	\citep{seidelPredictingTreeSpecies2021c} demonstrated $86\%$ accuracy for seven species from 2D renderings, and \citep{allenTreeSpeciesClassification2023a} reached $80.6\%$ via joint multi-viewpoint classification.
	The FOR-species20K benchmark of \citep{pulitiBenchmarkingTreeSpecies2025a} confirmed that the 2D model DetailView ($79.5\%$ OA) still outperformed 3D methods ($72\%$ average OA), highlighting that projection-based methods benefit from CNN maturity at the cost of discarding explicit 3D geometry.
	
	\smallskip
	\noindent\textbf{Direct 3D methods.}
	Recent architectures have operated on raw coordinates.
	\citep{liuTreeSpeciesClassification2021d} proposed LayerNet, which partitions point clouds into overlapping Euclidean layers, reaching $92.5\%$ accuracy.
	\citep{xiSeeForestTrees2020b} benchmarked 13 classifiers across nine species and found PointNet++ the strongest (mIoU $0.906$).
	\citep{liuMultifeatureFusionNetwork2025} introduced MFFTC-Net---combining boundary-driven sampling with an umbrella-RepSurf module---achieving state-of-the-art results on STPCTLS.
	\citep{WANG2024114456} proposed a cross-branch transformer for sparse multispectral ALS ($9$~pts/m$^2$), attaining $83.1$--$95.3\%$ OA across species-to-leaf-type granularities.
	
	\smallskip
	\noindent\textbf{DGCNN on TLS/ALS tree data.}
	DGCNN has been widely adopted for tree species classification across various acquisition platforms.
	On ALS, \citep{taher2026multispectral} obtained $87.0\%$ OA on high-density multispectral data with ensemble voting, marginally below Point Transformer, while \citep{pulitiBenchmarkingTreeSpecies2025a} reported substantially lower accuracy on proximal scans, attributing the drop to aggressive subsampling.
	On TLS, \citep{liu2022treespecies_mls} achieved $96.47\%$ on mobile laser scanning, confirming EdgeConv's effectiveness for high-density data, whereas \citep{liuMultifeatureFusionNetwork2025} placed DGCNN behind MFFTC-Net on STPCTLS.
	\citep{kandgcnn} proposed liteDGCNN, substituting linear layers with STFT-KAN operations to match baseline accuracy with $87\%$ fewer parameters.
	Across all variants, the same edge encoding is applied regardless of the neighborhood scale, a limitation that MS-DGCNN++ addresses through scale-dependent encoding (Section~\ref{sec:robustness}).
	
	%---------------------------------------------------------------
	\subsection{Multi-Scale and Adaptive Neighbourhood Methods}
	\label{subsec:multiscale_methods}
	%---------------------------------------------------------------
	
	Capturing the structure at multiple spatial scales is a central challenge in point cloud representation learning.
	We survey the main strategies below to situate our contribution to the literature.
	
	\smallskip
	\noindent\textbf{Multi-radius grouping.}
	PointNet++~\citep{qi2017pointnet++} introduced Multi-Scale Grouping (MSG), querying concentric ball radii at each centroid and concatenating the resulting features.
	PointNeXt~\citep{qian2022pointnext} revisited this framework with inverted residual bottlenecks and improved training recipes, achieving strong results without architectural novelty improvements.
	RS-CNN~\citep{rscnn} enriches grouping by explicitly encoding the distance, angle, and direction; however, the relation mapping is shared across all levels without adapting to the noise profile of each scale.
	In all cases, the grouping radius varied, whereas the edge encoding remained scale-agnostic.
	
	\smallskip
	\noindent\textbf{Dilated and continuous convolutions.}
	RandLA-Net~\citep{randlanet} stacks dilated residual blocks that double the receptive field at each stage via attentive $k$-NN aggregation, enabling the single-pass processing of $10^6$ points with a uniform attention mechanism across all dilation levels.
	PointConv~\citep{pointconv} models convolution as a continuous function of local coordinates, using an MLP to implicitly learn density-dependent weights, although this addresses density variation through post-hoc reweighting rather than through explicit scale-dependent edge construction.
	CurveNet~\citep{curvenet} aggregates features along learned curves, with an effective scale governed by the curve length rather than an explicit radius.
	
	\smallskip
	\noindent\textbf{Surface-aware, MLP-based, and attention methods.}
	RepSurf~\citep{repsurf} augments point features with triangular and umbrella surface representations that encode curvature at the neighborhood level.
	PointMLP~\citep{pointmlp} demonstrated that a residual MLP with geometric affine transformations achieves competitive accuracy, with multiscale behavior emerging from hierarchical sampling rather than encoding variation.
	Point Transformer~V2~\citep{wu2024point} introduces grouped vector self-attention with partition-based pooling across non-overlapping spatial groups.
	Point-M2AE~\citep{zhang2022point} constructs a three-level pyramid encoder with multi-scale masking and a global-to-local decoder that models cross-scale dependencies during the pretraining.
	
	\smallskip
	\noindent\textbf{Common limitation.}
	Despite this diversity, a fundamental gap persists: \emph{none of these methods adapt the edge or feature encoding to the spatial scale of the neighborhood }.
	Ball queries, dilated $k$-NN, continuous convolutions, and grouped attention all apply the same encoding function at each scale.
	In forest LiDAR, where trunk spacing (${\sim}5$\,mm) and canopy spacing (${\sim}300$\,mm) coexist, this forces certain neighborhood levels into a suboptimal noise regime in which raw edge magnitudes reflect density variation rather than intrinsic geometry.
	MS-DGCNN++ closes this gap by assigning a scale-dependent encoding to each neighborhood level, derived from the noise sensitivity analysis in Section~\ref{sec:robustness}.
	
	%---------------------------------------------------------------
	\subsection{Dynamic Graph Convolutional Neural Network (DGCNN)}
	\label{subsec:dgcnn}
	%---------------------------------------------------------------
	
	Having established that no existing multi-scale method adapts its encoding to the neighborhood scale, we now review the specific architecture upon which MS-DGCNN++ is built.
	
	\subsubsection{Edge Convolution (EdgeConv)}
	
	Consider a point cloud $X = \{\mathbf{x}_1, \ldots, \mathbf{x}_n\} \subseteq \mathbb{R}^D$.
	The DGCNN~\citep{wang2019dynamic} constructs a directed graph $\mathcal{G} = (\mathcal{V}, \mathcal{E})$ where each vertex connects to its $k$ nearest neighbors in the current feature space.
	For each edge $(i,j) \in \mathcal{E}$, an edge feature is computed as
	\begin{equation}
		\mathbf{e}_{ij} = h_{\boldsymbol{\Theta}}(\mathbf{x}_i, \mathbf{x}_j - \mathbf{x}_i),
	\end{equation}
	where $h_{\boldsymbol{\Theta}}$ is a shared MLP that processes both the center point and displacement vector.
	Symmetric aggregation (typically max-pooling) combines the edge features per vertex as follows:
	\begin{equation}
		\mathbf{x}'_{i} = \max_{j: (i,j)\in \mathcal{E}} h_{\boldsymbol{\Theta}}(\mathbf{x}_i, \mathbf{x}_j - \mathbf{x}_i).
	\end{equation}
	
	\subsubsection{Adaptive Graph Reconstruction}
	
	A key distinction from static graph networks is DGCNN's dynamic graph updating of DGCNN.
	After each layer, $k$-NN relationships are recomputed in the transformed feature space, allowing spatially distant points to connect through the learned semantic affinity.
	This enables a progression from local geometric similarity in the early layers to global semantic relationships in the deeper layers.
	
	\subsubsection{Multi-Scale DGCNN (MS-DGCNN)}
	
	MS-DGCNN~\citep{zhaiMultiScaleDynamicGraph2020c} employs farthest-point sampling (FPS) to select 512 representative points, and then applies three parallel EdgeConv branches at $k{=}20, 30, 40$.
	Multi-scale features are concatenated and processed through shared MLPs ($448 \to 512 \to 1024$), global max pooling, and classification layers ($1024 \to 512 \to 256 \to \text{output}$). While MS-DGCNN represents a natural first step toward multi-scale graph processing, it suffers from the four deficiencies analyzed in Section~\ref{sec:intro}: redundant encoding, no cross-scale interaction, no theoretical grounding, and density sensitivity.
	These limitations motivated the hierarchical architecture and scale-dependent encoding proposed in this study.
	
	\section{Method: MS-DGCNN++}
	\label{sec:method}
	
	MS-DGCNN++ operates through two stages: (1)hierarchical multi-scale feature initialization on raw 3D coordinates with scale-dependent edge encodings and (2)dynamic graph convolution in the learned feature space.
	The complete architecture is illustrated in Figure~\ref{fig:complete_model}.
	
	\subsection{Multi-Scale $k$-NN Graph Construction}
	\label{sec:graph_construction}
	
	Given a point cloud $\mathbf{X} \in \mathbb{R}^{B \times D \times N}$ 
	($B$: batch size, $D{=}3$, $N$: number of points), we compute a single 
	pairwise distance matrix $\mathbf{D}_{ij} = -\|\mathbf{x}_i - 
	\mathbf{x}_j\|_2^2$ and extract $k$-nearest-neighbour indices at two 
	scales simultaneously: the local scale ($k_1{=}5$), which captures 
	fine-grained surface geometry such as leaf and branch detail, and the 
	intermediate scale ($k_2{=}20$), which captures broader structural 
	patterns, such as crown architecture and branching angles.
	A third parameter ($k_3{=}30$) governs the dynamic graph in the learned 
	feature space (Section~\ref{sec:learned_space}).
	These defaults were selected via the sensitivity analysis in 
	Section~\ref{sec:kscale_sensitivity}.
	
	\subsection{Scale-Dependent Edge Encoding}
	\label{sec:edge_encoding}
	
	The key design principle of MS-DGCNN++ is that each scale receives a different type of edge encoding.
	For point $\mathbf{x}_i$ and neighbor $\mathbf{x}_j$, we define the relative displacement $\mathbf{r}_{ij} = \mathbf{x}_j - \mathbf{x}_i$ and the normalized direction $\hat{\mathbf{r}}_{ij} = \mathbf{r}_{ij} / (\|\mathbf{r}_{ij}\| + \epsilon)$, with $\epsilon = 10^{-8}$.
	
	\paragraph{Local scale ($k_1$).}
	The edge feature concatenates the raw displacement with the center point coordinates as follows:
	\begin{equation}
		\mathbf{F}_1 = \bigl[\,\mathbf{r}_{ij} \;\|\; \mathbf{x}_i\,\bigr] \in \mathbb{R}^{6}.
	\end{equation}
	At this scale, the neighbor distances are small, and the signal-to-noise ratio for directional estimation is low (Section~\ref{sec:robustness}); the raw encoding preserves the fine-grained metric geometry without injecting noisy directional estimates.
	
	\paragraph{Intermediate scale ($k_2$).}
	The edge feature additionally includes the normalized direction as follows:
	\begin{equation}
		\mathbf{F}_2 = \bigl[\,\mathbf{r}_{ij} \;\|\; \hat{\mathbf{r}}_{ij} \;\|\; \mathbf{x}_i\,\bigr] \in \mathbb{R}^{9}.
	\end{equation}
	At this scale, neighbor distances are larger, and the normalized direction provides a density-invariant angular descriptor with MSE decaying as $\mathcal{O}(1/s^2)$ (Theorem~\ref{thm:relative}).
	The raw displacement is retained to preserve the absolute distance information, exploiting the complementary noise profiles of the two representations (Proposition~\ref{prop:hybrid}).
	
	\subsection{Hierarchical Feature Fusion}
	\label{sec:fusion}
	
	Each scale's edge features are processed by a dedicated $1 \times 1$ convolution ($\phi_1 : \mathbb{R}^6 \to \mathbb{R}^{64}$, $\phi_2 : \mathbb{R}^9 \to \mathbb{R}^{64}$) followed by batch normalization and LeakyReLU, then aggregated over neighbors via channel-wise max-pooling:
	\begin{equation}
		\mathbf{G}_s = \max_{j \in \mathcal{N}_s(i)} \;\phi_s(\mathbf{F}_s) \in \mathbb{R}^{B \times 64 \times N}, \quad s \in \{1, 2\}.
	\end{equation}
	The two per-point feature vectors are fused by concatenation and projection through a learned $1 \times 1$ convolution:
	\begin{equation}
		\mathbf{Z} = \psi\bigl([\mathbf{G}_1 \;\|\; \mathbf{G}_2]\bigr), \quad \psi : \mathbb{R}^{128} \to \mathbb{R}^{64}.
	\end{equation}
	Alternative fusion operators (element-wise addition, attention, gated, and squeeze-and-excitation) are compared in Section~\ref{sec:component_ablation}.
	
	\subsection{Dynamic Graph Convolution in Learned Feature Space}
	\label{sec:learned_space}
	
	After fusion, three successive EdgeConv layers operated in the learned feature space using $k_3{=}30$.
	At each layer $\ell \in \{2, 3, 4\}$, the $k_3$-NN graph is \emph{recomputed} from the current features, allowing the topology to adapt as the representations evolve:
	\begin{equation}
		\mathbf{x}'_i = \max_{j \in \mathcal{N}_{k_3}^{(\ell)}(i)} \;\phi_\ell\bigl([\,\mathbf{x}_j - \mathbf{x}_i \;\|\; \mathbf{x}_i\,]\bigr).
	\end{equation}
	These layers used standard DGCNN edge encoding (raw relative features concatenated with center features) \emph{without} normalization.
	In the learned space, the Euclidean distance reflects semantic similarity rather than geometric proximity; therefore, the density-dependent magnitude artifacts that motivate normalization in the coordinate space do not arise.
	
	The output channels were 64, 128, and 256 for layers 2, 3, and 4, respectively.
	All four per-point feature vectors were concatenated via skip connections as follows:
	\begin{equation}
		\mathbf{H} = [\,\mathbf{Z} \;\|\; \mathbf{x}_2 \;\|\; \mathbf{x}_3 \;\|\; \mathbf{x}_4\,] \in \mathbb{R}^{B \times 512 \times N}.
	\end{equation}
	
	\subsection{Global Aggregation and Classification}
	\label{sec:classification}
	
	The concatenated features are projected to 1024 channels via a $1 \times 1$ convolution, then aggregated by dual global pooling as follows:
	\begin{equation}
		\mathbf{g} = \bigl[\,\text{MaxPool}(\mathbf{H}') \;\|\; \text{AvgPool}(\mathbf{H}')\,\bigr] \in \mathbb{R}^{B \times 2048}.
	\end{equation}
	A classification head of two fully connected layers (512, 256 units) with batch normalization, LeakyReLU, and dropout ($p{=}0.5$) produced the final class logits.
	
	\begin{figure*}[!tbp]
		\centering
		\begin{subfigure}[b]{\textwidth}
			\centering
			\includegraphics[scale=0.35]{ms_dgcnn2_arch1.pdf}
			\caption{Multi-Scale Fusion Module: two-scale feature extraction with scale-dependent edge encoding, followed by learned fusion.}
			\label{fig:fusion_module}
		\end{subfigure}

		\begin{subfigure}[b]{\textwidth}
			\centering
			\includegraphics[scale=0.35]{ms_dgcnn2_arch2.pdf}
			\caption{Dynamic Graph Convolution Module: successive EdgeConv layers in learned feature space with $k_3$-NN graph recomputation.}
			\label{fig:dgcnn_module}
		\end{subfigure}

		\begin{subfigure}[b]{\textwidth}
			\centering
			\includegraphics[scale=0.35]{ms_dgcnn2_arch3.pdf}
			\caption{Classification Module: dual global pooling and MLP classifier.}
			\label{fig:classification_module}
		\end{subfigure}
		
		\caption{Complete MS-DGCNN++ architecture. Stage1 (a) extracts and fuses multi-scale features on raw coordinates using scale-dependent encoding. Stage2 (b) refines the features through dynamic graph convolution in the learned feature space. Stage~3 (c) produces class predictions.}
		\label{fig:complete_model}
	\end{figure*}

	\section{Robustness of Scale-Dependent Edge Encoding}
	\label{sec:robustness}
	
	In ALS/TLS forest acquisitions, point density varies by orders of magnitude within a single scene: TLS produces spacings of $1$--$5$\,mm on nearby trunks, while ALS yields $0.5$--$2$\,m in the upper canopy.
	Because graph-based architectures encode geometry through edge features $\rvec = q - p$, this heterogeneity causes raw edge magnitudes to reflect point spacing rather than intrinsic morphology.
	MS-DGCNN++ addresses this issue via scale-dependent encoding.
	This section proves that the asymmetric design is noise-optimal.
	
	\begin{remark}[Unit convention]
		In all experiments, the point clouds were normalized to a unit sphere prior to processing (Section~\ref{sec:preprocessing}).
		The noise model below is stated in physical units (mm) for interpretability; however, because the unit-sphere normalization is a global affine transform, the signal-to-noise ratio $\SNR = s/\sigma$ and all derived thresholds are invariant to this rescaling.
	\end{remark}
	
	\subsection{Noise Model and Edge Encodings}
	\label{sec:setup}
	
	Let $p,q\in\R^D$ ($D=3$) be true positions with noisy measurements
	$\tilde p = p + \eps_p$, $\tilde q = q + \eps_q$, where $\eps_p,\eps_q\sim\mathcal{N}(0,\sigma^2 I_D)$ independently.
	The true and noisy relative vectors are $\rvec = q-p$ and $\tilde\rvec = \rvec + \eps$, respectively, where $\eps = \eps_q-\eps_p\sim\mathcal{N}(0,2\sigma^2 I_D)$.
	We write $s = \norm{\rvec}$ and $\SNR = s/\sigma$.
	
	\begin{definition}[Edge encodings]
		\label{def:edge}
		\begin{enumerate}[label=(\roman*)]
			\item \emph{Raw}: $\phi_{\mathrm{raw}}(p,q) = q - p$.
			\item \emph{Normalised}: $\phi_{\mathrm{dir}}(p,q) = (q-p)/\norm{q-p}$.
			\item \emph{Hybrid} (intermediate scale $k_2$):
			$\phi_{\mathrm{hyb}}(p,q) = [\,q{-}p\;\|\;\tfrac{q-p}{\norm{q-p}}\;\|\;p\,] \in \R^{3D}$.
			\item \emph{Local} (scale $k_1$):
			$\phi_{\mathrm{loc}}(p,q) = [\,q{-}p\;\|\;p\,] \in \R^{2D}$.
		\end{enumerate}
	\end{definition}
	
	\subsection{MSE of the Raw Relative Vector}
	\label{sec:raw_mse}
	
	\begin{proposition}[MSE of $\phi_{\mathrm{raw}}$]
		\label{prop:unnorm}
		$\E\bigl[\norm{\tilde\rvec - \rvec}^2\bigr] = 2D\,\sigma^2$,
		independent of $s = \norm{\rvec}$.
	\end{proposition}
	
	\begin{proof}
		Since $\tilde\rvec - \rvec = \eps \sim \mathcal{N}(0, 2\sigma^2 I_D)$,
		$\E[\norm{\eps}^2] = \tr(2\sigma^2 I_D) = 2D\sigma^2$.
	\end{proof}
	
	The constant MSE means that the \emph{relative} corruption $\norm{\eps}/s$ diverges as $s \to 0$ and vanishes as $s \to \infty$.
	For a TLS scan with $\sigma = 10$\,mm, the RMS noise magnitude is $\sqrt{2D}\,\sigma \approx 24.5$\,mm.
	At the local scale ($k_1{=}5$, typical $s \approx 20$\,mm), the relative error is ${\sim}1.22$: the noise exceeds the signal.
	At the intermediate scale ($k_2{=}20$, typical $s \approx 300$\,mm), the relative error dropped to ${\sim}0.08$.
	
	However, when the density is heterogeneous, two intermediate-scale neighbors at similar branching angles but different spacings ($s_1{=}100$\,mm vs.\ $s_2{=}800$\,mm) produce raw vectors differing eightfold in magnitude.
	The larger magnitude dominates max-pooling not because it is geometrically more informative but because the spacing is wider.
	A normalized representation eliminates this bias.
	
	\subsection{MSE of the Normalised Direction}
	\label{sec:norm_mse}
	
	\begin{theorem}[MSE of $\phi_{\mathrm{dir}}$]
		\label{thm:relative}
		For any $\rvec\neq0$ with $\norm{\rvec} = s$,
		\[
		\E\bigl[\norm{\tilde\uvec - \uvec}^2\bigr] = \frac{2(D-1)\sigma^2}{s^2} + R(s,\sigma),
		\]
		where $\uvec = \rvec/s$, $\tilde\uvec = \tilde\rvec/\norm{\tilde\rvec}$, and $\abs{R(s,\sigma)} \le C_D\,\sigma^4/s^4$ for $\SNR > \alpha_D$.
	\end{theorem}
	
	\begin{proof}
		Apply the multivariate delta method to $f(x)=x/\norm{x}$ with Jacobian
		$J_f(x) = \norm{x}^{-1} P_{\hat{x}}$,
		where $P_{\hat{x}} = I_D - \hat x\hat x^{\mathsf T}$ is the orthogonal projector perpendicular to $\hat x$, satisfying $P_{\hat{x}}^2 = P_{\hat{x}}$ and $\tr(P_{\hat{x}}) = D-1$.
		
		First-order expansion around $\rvec$:
		$\tilde\uvec - \uvec = s^{-1} P_{\uvec}\,\eps + \mathcal{R}$,
		where $\E[\norm{\mathcal{R}}^2] \le C_D\,\sigma^4/s^4$ for $\SNR > \alpha_D$, according to a standard Taylor remainder bound.
		For the leading term:
		\[
		\E\!\left[\left\lVert\frac{1}{s}\,P_{\uvec}\,\eps\right\rVert^2\right] = \frac{2\sigma^2}{s^2}\,\tr(P_{\uvec}) = \frac{2(D-1)\sigma^2}{s^2}.
		\]
		The cross term is absorbed into the remainder by the Cauchy--Schwarz.
	\end{proof}
	
	For $D=3$, the leading term is $4\sigma^2/s^2$, which is negligible when $\SNR \gtrsim 5$ (i.e., $\sigma = 10$\,mm and $s \ge 50$\,mm).
	
	\subsection{Normalisation Threshold}
	\label{sec:threshold}
	
	\begin{corollary}
		\label{cor:threshold}
		The normalised direction has lower leading-order MSE than the raw vector when
		$s > \sqrt{(D-1)/D}\;\sigma$.
		For $D=3$: $s > \sqrt{2/3}\,\sigma \approx 0.816\,\sigma$.
	\end{corollary}
	
	\begin{proof}
		Set $2(D-1)\sigma^2/s^2 < 2D\,\sigma^2$, yielding $s^2 > \tfrac{D-1}{D}\,\sigma^2$.
	\end{proof}
	
	\begin{remark}
		The raw and normalized vectors live in $\R^D$ and $S^{D-1}$ respectively, so direct MSE comparison is a proxy.
		The operationally relevant distinction is the \emph{scaling behavior }: constant for raw versus $\mathcal{O}(1/s^2)$ for normalized data.
		Downstream convolutions rescale features; what matters is whether the noise systematically degrades or improves with the neighborhood radius.
	\end{remark}
	
	\subsection{Complementary Information in the Hybrid Encoding}
	\label{sec:hybrid}
	
	MS-DGCNN++ concatenates both raw and normalized vectors (plus the absolute position) at the intermediate scale.
	This is justified by the complementary noise profiles.
	
	\begin{proposition}[Complementary information content]
		\label{prop:hybrid}
		Decompose $\rvec = s\,\uvec$ with $s \in \R_{>0}$ and $\uvec \in S^{D-1}$.
		Under the noise model of Theorem~\ref{thm:relative},
		\begin{enumerate}[label=(\alph*)]
			\item The noisy distance $\tilde s = \norm{\tilde\rvec}$ satisfies
			$\E[(\tilde s - s)^2] = 2D\,\sigma^2 - 2(D{-}1)\sigma^4/s^2 + O(\sigma^6/s^4) \to 2D\,\sigma^2$ as $\SNR\to\infty$.
			\item The directional MSE is $2(D{-}1)\sigma^2/s^2 + O(\sigma^4/s^4)$, \emph{decreasing} with $s$.
		\end{enumerate}
		Hence, the direction becomes more reliable at larger distances, while the magnitude retains the absolute scale that the direction discards.
	\end{proposition}
	
	\begin{proof}
		For~(a): $\tilde s^2 = s^2 + 2\rvec^{\mathsf T}\eps + \norm{\eps}^2$, so $\E[\tilde s^2] = s^2 + 2D\sigma^2$.
		Apply the delta method to $g(y) = \sqrt{y}$: $\E[\tilde s] = s + D\sigma^2/s + O(\sigma^4/s^3)$, and the variance follows from $\Var(\tilde s) = \E[\tilde s^2] - (\E[\tilde s])^2$.
		Part~(b) restates Theorem~\ref{thm:relative}.
	\end{proof}
	
	\subsection{Implications for Max-Pooling and Scale-Dependent Design}
	\label{sec:maxpool}
	
	MS-DGCNN++ aggregates edge features via channel-wise max-pooling, which adversely interacts with density-induced magnitude variation. Under raw encoding, sparse neighbors produce inflated activations that dominate pooling, regardless of geometric informativeness.
	The normalized channels project edges onto $S^{D-1}$ before convolution, so max-pooling selects the most angularly distinctive direction rather than the most distant point from the center.
	Thus, hybrid encoding combines distance-dependent information (raw channels) with density-invariant directional descriptors (normalized channels).
	
	\begin{remark}[Local scale omits normalisation]
		\label{rem:local}
		At $k_1{=}5$ with a typical $\sigma \approx 10$\,mm and $s \approx 20$\,mm, the directional MSE is $4\sigma^2/s^2 \approx 1$: the unit vector is noise-dominated. Omitting normalization here is a variance reduction strategy that preserves the fine-grained local geometry.
	\end{remark}
	
	\begin{remark}[Intermediate scale includes normalisation]
		\label{rem:branch}
		At $k_2{=}20$, the neighborhood spans regions of heterogeneous density, where raw magnitudes differ enormously, even for similar branching angles. Normalization isolates the angular structure (MSE $\propto 1/s^2$); retaining the raw vector alongside preserves the metric distance.
	\end{remark}
	
	%=============================================================================
	\section{Experimental Setup}
	\label{sec:exp}
	\subsection{Datasets}
	To evaluate the performance of MS-DGCNN++, we conducted various experiments on two LiDAR datasets for tree species classification: STPCTLS and HeliAL.
	\subsubsection{STPCTLS}
	
	For the evaluation of the TLS data, we utilized the terrestrial laser scanning dataset (STPCTLS) compiled by \citep{seidelPredictingTreeSpecies2021c}, which comprises high-resolution point cloud measurements acquired from both artificial and natural forest environments in Germany and the United States. The dataset encompasses seven morphologically distinct tree species---beech, red oak, ash, oak, Douglas fir, spruce, and pine---selected on the basis of their pronounced inter-species morphological similarities alongside substantial intra-species variability arising from heterogeneous growth conditions. These characteristics collectively increase the complexity of the classification task, necessitating robust and discriminative feature extraction capabilities. Notably, this dataset provides only raw XYZ geometric coordinates as the input features. Given the relatively limited number of available samples, a stratified $k$-fold cross-validation protocol is adopted to ensure reliable performance estimation while preserving class-proportion balance 
	across the folds; full experimental details are reported in Table~\ref{tab:stpctlc_summary}.
	
	\begin{table}[!tb]
		\centering
		\caption{STPCTLC dataset summary: number of samples per tree species. 5-fold 
			stratified cross-validation is used (no held-out test set): 691 total samples, 
			5 folds, $\sim$552 training samples (80\%) and $\sim$138 validation samples 
			(20\%) per fold, with stratification to preserve class proportions.}
		\label{tab:stpctlc_summary}
		\begin{tabular}{p{2cm}p{2cm}p{2cm}}
			\toprule
			Species & Samples & \% \\
			\midrule
			Buche      & 164 & 23.7 \\
			Douglasie  & 183 & 26.5 \\
			Eiche      &  22 &  3.2 \\
			Esche      &  39 &  5.6 \\
			Fichte     & 158 & 22.9 \\
			Kiefer     &  25 &  3.6 \\
			Roteiche   & 100 & 14.5 \\
			\midrule
			\textbf{Total} & \textbf{691} & \textbf{100.0} \\
			\bottomrule
		\end{tabular}
	\end{table}
	
	\subsubsection{HeliALS Dataset}
	\label{sec:ms-als-species}
	To evaluate our method on ALS data, we used HeliALS, a version of the MS-ALS-SPECIES benchmark \citep{taher2026multispectral} for tree species classification. The dataset comprises 6,326 individual tree samples spanning nine species: pine, spruce, birch, maple, aspen, rowan, oak, linden, and alder, collected over Espoonlahti, Finland. Point clouds were acquired using a helicopter-borne three-wavelength active laser scanning system (HeliALS) at 100~m above ground level, yielding high-density scans exceeding 1,000~points/m$^2$. Table~\ref{tab:helials_summary} presents the per-species sample distributions across the training, validation, and test sets. It should be noted that the test set is predefined by the dataset authors.
	
	\begin{table}[!tb]
		\centering
		\caption{HeliALS dataset summary: sample distribution per tree species across 
			train/validation/test splits. Only raw XYZ geometric coordinates are employed 
			as input features, excluding any additional spectral or intensity attributes.}
		\label{tab:helials_summary}
		\begin{tabular}{lcccccc}
			\toprule
			& \multicolumn{2}{c}{Train} & \multicolumn{2}{c}{Validation} & \multicolumn{2}{c}{Test} \\
			\cmidrule(lr){2-3} \cmidrule(lr){4-5} \cmidrule(lr){6-7}
			Species & $n$ & \% & $n$ & \% & $n$ & \% \\
			\midrule
			Pine & 270 & 28.2 & 30 & 28.0 & 2047 & 38.9 \\
			Spruce & 135 & 14.1 & 15 & 14.0 & 528 & 10.0 \\
			Birch & 270 & 28.2 & 30 & 28.0 & 1258 & 23.9 \\
			Maple & 90 & 9.4 & 10 & 9.3 & 587 & 11.2 \\
			Aspen & 90 & 9.4 & 10 & 9.3 & 376 & 7.1 \\
			Rowan & 36 & 3.8 & 4 & 3.7 & 171 & 3.3 \\
			Oak & 18 & 1.9 & 2 & 1.9 & 49 & 0.9 \\
			Linden & 36 & 3.8 & 4 & 3.7 & 129 & 2.5 \\
			Alder & 13 & 1.4 & 2 & 1.9 & 116 & 2.2 \\
			\midrule
			\textbf{Total} & \textbf{958} & \textbf{15.1} & \textbf{107} & \textbf{1.7} & \textbf{5261} & \textbf{83.2} \\
			\bottomrule
		\end{tabular}
	\end{table}
	
	\subsection{Data Preprocessing}
	\label{sec:preprocessing}
	
	We standardized point cloud density across all datasets through farthest-point sampling (FPS)~\citep{eldar1997farthest}, downsampling to 1,024 points for STPCTLS and 2,048 points for HeliALS.
	FPS iteratively selects the point furthest from all previously selected points, preserving the geometric structure by maximizing inter-point distances, which is particularly important for tree classification, where fine-grained morphological details, such as branching patterns and crown architecture, must be retained despite significant reductions in point count.
	
	Each point cloud was then normalized to a unit sphere by translating all points to the centroid and scaling them by the maximum distance from the origin.
	This eliminates scale variation across samples, ensuring that the classification is not confounded by differences in absolute tree size or scan range.
	
	For online data augmentation, each point cloud was randomly rotated about the vertical axis, preserving the natural upright orientation of the tree structures while introducing variability in the horizontal orientation to improve the robustness of the differing scanner positions.
	
	In addition to clean-data evaluation, several experiments apply controlled perturbations to the input at training and/or test time (Figure~\ref{fig:transformations}): Gaussian noise injection to simulate sensor measurement error, upper-canopy point dropout to simulate occlusion and density heterogeneity, and outlier injection to simulate spurious returns.
	These transformations are described in detail in the corresponding ablation and robustness sections of this paper.
	
	\begin{figure}[!tb]
		\centering
		\begin{subfigure}{0.5\linewidth}
			\centering
			\includegraphics[width=\linewidth]{transformation_noise.pdf}
			\caption{Gaussian noise injection.}
			\label{fig:trans_noise}
		\end{subfigure}
		
		\begin{subfigure}{0.5\linewidth}
			\centering
			\includegraphics[width=\linewidth]{transformation_dropout.pdf}
			\caption{Upper-canopy point dropout.}
			\label{fig:trans_dropout}
		\end{subfigure}
		
		\begin{subfigure}{0.5\linewidth}
			\centering
			\includegraphics[width=\linewidth]{transformation_outlier.pdf}
			\caption{Outlier injection.}
			\label{fig:trans_outlier}
		\end{subfigure}
		
		\caption{Data transformations applied during ablation and robustness evaluation. Depending on the experiment, perturbations are applied at training time, test time, or both.}
		\label{fig:transformations}
	\end{figure}
	
	\subsection{Common Hyperparameters and Training Environment}
	
	A standardized hyperparameter configuration was maintained across all experiments in this study to ensure consistency, reproducibility, and fair comparison. The network was trained for 250 epochs using the AdamW optimizer with a learning rate of $5 \times 10^{-4}$ and a weight decay of $1 \times 10^{-4}$. A cosine annealing learning rate scheduler (CosLR) was employed with a warm-up phase of 10 epochs, and gradient clipping was applied with a maximum norm of 10 to stabilize the training. The batch size was set to 16 samples. For all baseline models, the original architectural parameters were preserved, as defined by their respective authors, to ensure the faithful reproduction of each method. For models falling under the same DGCNN-based category, category-specific parameters were additionally unified across architectures, notably adopting an embedding dimension of 1024 with max pooling as the global aggregation method. Regularization was enforced using a dropout probability of 0.5, and LeakyReLU activation functions were configured with a negative slope of 0.2.
	All experiments were conducted using Open3D for data preprocessing and PyTorch for model development and training, with CUDA 11.8 for GPU-acceleration. The performance metrics were computed using Scikit-learn. The experimental platform consisted of an NVIDIA GeForce RTX 4060 Ti GPU (16\,GB VRAM), 32\,GB RAM, and an AMD Ryzen 7 5700X 8-core processor, running Ubuntu 22.04 LTS.
	
	\section{Ablation Studies and Robustness Evaluation}
	\label{sec:ablation}
	\subsection{Per-Scale Encoding Ablation (Experiment~1)}
	\label{sec:exp1}
	
	We compared four encoding strategies: (a)~Raw-Only~(RO), raw vectors at both scales; (b)~Hybrid-Everywhere~(HE), normalized directions appended at both scales; (c)~Default Asymmetric~(DA), raw at local and hybrid at intermediate (proposed); and (d)~Reversed Asymmetric~(RA), which swaps this assignment.
	
	All normalization-equipped variants~(b--d) outperform the raw-only variant ~(a) by $+4.2$--$6.1\%$ in terms of the Overall Accuracy(OA) (Table~\ref{tab:exp1_results}), confirming that normalized directional features provide a consistent benefit.
	Variant~(b) achieves the highest OA ($90.30\%$), whereas ~(c) attains the best per-class F1 on the hardest minority species---Eiche ($0.756$ vs.\ $0.704$) and Esche ($0.695$ vs.\ $0.528$)---consistent with the theoretical rationale: species distinguished by crown architecture benefit most from density-invariant directional descriptors at the intermediate scale.
	The reversed variant~(d) underperforms both~(b) and~(c), confirming that the \emph{assignment} of encoding to scale matters: normalized directions help where inter-point distances support reliable estimation (Remark~\ref{rem:branch}) but inject noise at the local scale where the SNR is low (Remark~\ref{rem:local}).
	
	\begin{table*}
		\scriptsize
		\caption{Per-scale encoding ablation on STPCTLS (5-fold CV). RO: Raw-Only; HE: Hybrid-Everywhere; DA: Default Asymmetric (proposed); RA: Reversed Asymmetric. The per-class columns report the mean F1 scores. Bold indicates best per column.}
		\label{tab:exp1_results}
		\resizebox{\textwidth}{!}{\begin{tabular}{llllllllll}
				\toprule
				& OA $\pm$ std & mAcc $\pm$ std & Buche & Douglasie & Eiche & Esche & Fichte & Kiefer & Roteiche \\
				\midrule
				(a) RO & $84.23 \pm 4.07$ & $77.67 \pm 4.55$ & 0.889 & 0.879 & 0.659 & 0.586 & 0.844 & 0.673 & 0.870 \\
				(b) HE & \textbf{$90.30 \pm 1.42$} & \textbf{$84.60 \pm 3.94$} & 0.903 & \textbf{0.950} & 0.704 & 0.528 & \textbf{0.946} & \textbf{0.861} & \textbf{0.936} \\
				(c) DA & $88.86 \pm 2.31$ & $84.27 \pm 7.42$ & \textbf{0.915} & 0.906 & \textbf{0.756} & \textbf{0.695} & 0.896 & 0.815 & 0.906 \\
				(d) RA & $88.42 \pm 2.37$ & $81.83 \pm 3.98$ & 0.904 & 0.917 & 0.709 & 0.613 & 0.910 & 0.784 & 0.905 \\
				\bottomrule
		\end{tabular}}
	\end{table*}

	\subsection{Controlled Density Dropout (Experiment~2)}
	\label{sec:exp2}
	
	We simulated upper-canopy thinning by randomly removing points above the median height at retention rates $r \in \{100, 75, 50, 25, 10, 5\}\%$, training separate models at each rate.
	The secondary axis in Figure~\ref{fig:degradation_curves} tracks the mean $k_2$-neighbor distance, which increases monotonically, confirming that thinning inflates $\|\mathbf{r}_{ij}\|$.
	
	The default variant~(c) maintains the highest OA at mid-range retention ($r{=}50$--$10\%$), overtaking~(b) despite starting lower at full density.
	Raw-only~(a) drops only $3.6\%$ (partly a floor effect from its lower starting point), while~(c) drops $4.2\%$,~(b) drops $5.2\%$, and~(d) degrades most at $5.8\%$.
	At $r{=}5\%$,~(b) retains a slight edge over~(c) ($85.1\%$ vs.\ $84.7\%$), but its steeper slope suggests greater vulnerability to further reduction.
	The faster degradation of~(b) is consistent with local normalization amplifying noise as $\text{SNR}_1$ collapses under sparsity.

	\begin{figure}[!tb]
		\centering
		\begin{tabular}{c}
			\includegraphics[scale=0.25]{degradation_curves.pdf} \\
			
		\end{tabular}
		\caption{Accuracy vs.\ upper-canopy retention rate. Solid lines: OA for each variant (shaded bands: $\pm 1$ std over five folds). Dashed grey: mean $k_2$-neighbor distance (right axis), confirming that thinning inflates raw edge magnitudes. The default asymmetric variant~(c) achieves the highest OA at intermediate retention rates ($r{=}50$--$10\%$).}
		\label{fig:degradation_curves}
	\end{figure}

	\subsection{Synthetic Noise Sweep (Experiment~3)}
	\label{sec:exp3}
	
	Corollary~\ref{cor:threshold} predicts that the normalized directions have a lower MSE when $\text{SNR}_2 \gtrsim 1.22$.
	We added Gaussian noise $\mathcal{N}(0, \sigma^2 I_3)$ at ten levels ($\sigma \in [0.5, 150]$\,mm) and trained separate models at each level.
	
	At high SNR ($\sigma \le 5$\,mm), all variants perform within ${\sim}4\%$ (Figure~\ref{fig:accuracy_vs_snr}), consistent with theory.
	As noise increases, they diverge:~(c) achieves the best OA at 5 of 10 levels, dominating at $\sigma \ge 10$\,mm except at $\sigma{=}50$ and $100$\,mm where~(d) leads (Table~\ref{tab:exp3_full}).
	The dashed line marks the theoretical crossover at $\text{SNR}_2 \approx 1.22$; below it, the normalized variants gain a decisive advantage.
	Variant~(b) leads at low noise but deteriorates faster than~(c) beyond $\sigma{=}10$\,mm, again indicating that local normalisation becomes counterproductive at low $\text{SNR}_1$.
	
	\begin{figure}[!tb]
		\centering
		\includegraphics[scale=0.25]{accuracy_vs_snr.pdf}
		\caption{OA vs.\ empirical $\text{SNR}_2$ (log scale). The dashed red line marks the theoretical crossover at $\text{SNR}_2 \approx 1.22$ (Corollary~\ref{cor:threshold}). At a high SNR, all variants converge; below the crossover, normalized variants maintain a clear advantage. Shaded bands: $\pm 1$ std over 5~folds.}
		\label{fig:accuracy_vs_snr}
	\end{figure}

	\begin{table*}[!tb]
		\centering
		\caption{OA ($\%$) at ten Gaussian noise levels ($\sigma$ in mm). Bold indicates best per column. The default asymmetric variant~(c) dominates at moderate-to-high noise ($\sigma \ge 10$\,mm).}
		\label{tab:exp3_full}
		\resizebox{\textwidth}{!}{%
			\begin{tabular}{lcccccccccc}
				\toprule
				Variant & $\sigma$=0.5 & $\sigma$=1 & $\sigma$=2 & $\sigma$=5 & $\sigma$=10 & $\sigma$=20 & $\sigma$=50 & $\sigma$=75 & $\sigma$=100 & $\sigma$=150 \\
				\midrule
				(a) RO & 85.8$\pm$3.8 & 85.1$\pm$2.6 & 86.5$\pm$2.0 & 86.5$\pm$1.2 & 81.8$\pm$4.4 & 82.9$\pm$2.7 & 77.4$\pm$2.5 & 74.7$\pm$3.4 & 68.3$\pm$3.3 & 61.4$\pm$1.7 \\
				(b) HE & 88.0$\pm$3.3 & \textbf{91.2$\pm$2.0} & 90.6$\pm$1.9 & \textbf{90.0$\pm$4.5} & 83.6$\pm$2.1 & 81.9$\pm$3.4 & 76.7$\pm$1.8 & 72.8$\pm$3.7 & 69.5$\pm$3.0 & 61.4$\pm$3.8 \\
				(c) DE & \textbf{89.3$\pm$2.1} & 89.3$\pm$2.4 & 90.2$\pm$1.2 & 89.0$\pm$1.6 & \textbf{87.0$\pm$3.9} & \textbf{84.5$\pm$2.1} & 76.4$\pm$2.7 & \textbf{74.8$\pm$3.8} & 69.6$\pm$3.8 & \textbf{64.8$\pm$3.4} \\
				(d) RA & 89.0$\pm$3.5 & 90.7$\pm$2.0 & \textbf{90.9$\pm$2.2} & 87.3$\pm$3.4 & 86.5$\pm$2.9 & 79.0$\pm$7.5 & \textbf{78.9$\pm$2.3} & 72.1$\pm$1.7 & \textbf{71.3$\pm$2.5} & 62.8$\pm$2.5 \\
				\bottomrule
			\end{tabular}
		}
	\end{table*}
	
	\subsection{Max-Pooling Activation Provenance (Experiment~4)}
	\label{sec:exp4}
	
	To test the max-pooling bias predicted in Section~\ref{sec:maxpool}, we record which of the $k_2$ neighbors wins each channel-wise max-pool in the intermediate convolution $\phi_2$.
	Neighbours are split into \emph{near} (closer half) and \emph{far} (farther half); an unbiased aggregation should yield ${\sim}50\%$ each.
	
	In the raw-input model, far neighbors win $85.5\%$ at $r{=}100\%$ (Figure~\ref{fig:retention_bar_chart}), confirming that a larger $\|\mathbf{r}_{ij}\|$ dominates the aggregation.
	The normalized-input model was near-balanced ($52.4\%$/$47.6\%$).
	As retention decreases, the raw model's bias diminishes (to $77.6\%$ at $r{=}25\%$) because uniform thinning inflates all distances proportionally, compressing the near-to-far ratio.
	The normalized model remains within $\pm 2.5\%$ of chance at all retention rates.
	
	\begin{figure}[!tb]
		\centering
		\includegraphics[scale=0.07]{stacked_bar_chart.pdf}
		\caption{Max-pooling provenance at four retention rates. Each bar shows the fraction of channel-wise max-pool winners that are near (blue) vs.\ far (red) neighbors. The raw-input model exhibits a strong far-neighbor bias ($>77\%$), whereas the normalized-input model remains balanced at $50\%$ across all conditions. Dashed line: chance level.}
		\label{fig:retention_bar_chart}
	\end{figure}

	\subsection{Feature-Space Isotropy Analysis (Experiment~5)}
	\label{sec:exp5}
	
	We extract feature matrices at the local ($\mathbf{G}_1$), intermediate ($\mathbf{G}_2$), and fused ($\mathbf{Z}$) stages, computing the entropy-based effective rank $\text{erank} = \exp(-\sum_i p_i \log p_i)$ where $p_i = \sigma_i / \sum_j \sigma_j$.
	
	At the intermediate level, the default variant achieves an erank $31.05$ versus\ $16.94$ for raw-only---an $83\%$ increase (Figure~\ref{fig:vertical_stack})---confirming that normalization produces substantially more isotropic representations.
	At the local level, raw-only and default have comparable eranks (${\sim}15$), as expected.
	The hybrid-everywhere variant doubles the local erank to $30.53$; this correlates with higher clean-data accuracy (Exp.~1) but faster degradation under noise (Exp.~3), presenting a robustness--accuracy trade-off.
	The fused representation exceeded both constituents in all variants, confirming that fusion combines complementary rather than redundant information.
	All findings are stable across $r{=}100\%$ and $r{=}5\%$.
	
	\begin{figure}[!tb]
		\centering
		\includegraphics[scale=0.28]{effective_rank_bar.pdf}
		
		\vspace{0.5cm} 
		
		\includegraphics[scale=0.22]{singular_value_spectrum.pdf}
		
		\caption{Top: Effective rank by feature location and variant at $r{=}100\%$ and $r{=}5\%$. Normalization at the intermediate scale nearly doubled the effective rank compared to raw-only. Bottom: Normalized singular value spectra $\sigma_i/\sigma_1$ (log scale). Raw-only branch features exhibit steep decay (low isotropy), while the default and hybrid variants maintain a flatter spectrum, indicating more uniform use of the feature space.}
		\label{fig:vertical_stack}
	\end{figure}

	\subsection{Component Ablation: Single-Scale Baselines and Fusion}
	\label{sec:component_ablation}
	
	\paragraph{Single-scale baselines.}
	Six single-scale DGCNN variants ($k_1 \in \{5, 20, 30\}$, with/without normalization) were compared at $r{=}100\%$ and $r{=}5\%$ (Table~\ref{tab:ablation_results}).
	The raw DGCNN at $k_1{=}5$ nearly matches the multi-scale baseline at full density ($88.6\%$ vs.\ $88.9\%$) but collapses at $r{=}5\%$ ($78.4\%$, $-10.2\%$).
	Adding normalization reverses the $k$-dependence: DGCNN+Norm improves with larger $k$ ($84.7\% \to 89.0\%$), which is consistent with the SNR analysis.
	DGCNN+Norm at $k{=}30$ matches the baseline at $r{=}100\%$ but falls $2.3\%$ short at $r{=}5\%$, demonstrating that multi-scale fusion provides robustness beyond any single scale.
	
	\paragraph{Fusion strategies.}
	Addition ($85.7\%$) and attention ($83.5\%$) substantially underperformed concat\_conv ($88.9\%$).
	Gated fusion matches at $r{=}100\%$ ($89.0\%$) but degrades more at $r{=}5\%$ ($84.9\%$ vs.\ $86.7\%$).
	SE fusion achieves the best $r{=}5\%$ performance ($86.8\%$), marginally surpassing the default performance.
	Concat\_conv was retained for its overall balance, although SE fusion merits consideration for density-heterogeneous deployments.

	\begin{table*}[!tb]
		\centering
		\tiny
		\caption{Component ablation on STPCTLS (5-fold CV) at full density and extreme sparsity. The multi-scale default outperforms all single-scale baselines at $r{=}5\%$. Among the fusion strategies, concat\_conv and SE fusion offer the best accuracy--robustness balance. Bold indicates best within each section (within $0.5\%$ of maximum).}
		\label{tab:ablation_results}
		\resizebox{\textwidth}{!}{
			\begin{tabular}{l|cc|cc|cc}
				\toprule
				Model & \multicolumn{2}{c|}{OA (\%)} & \multicolumn{2}{c|}{mAcc (\%)} & \multicolumn{2}{c}{F1 (\%)} \\
				& r=100\% & r=5\% & r=100\% & r=5\% & r=100\% & r=5\% \\
				\midrule
				\multicolumn{7}{l}{\textbf{Baseline}} \\
				MS-DGCNN++ Default & \textbf{88.9$\pm$2.3} & \textbf{86.7$\pm$2.1} & 84.3$\pm$7.4 & \textbf{84.0$\pm$2.9} & 84.1$\pm$18.5 & \textbf{83.9$\pm$10.8} \\
				\midrule
				\multicolumn{7}{l}{\textbf{Single-Scale DGCNN}} \\
				DGCNN (k1=5, k=30) & \textbf{88.6$\pm$1.9} & 78.4$\pm$4.4 & 83.5$\pm$3.9 & 70.5$\pm$7.0 & 84.0$\pm$12.6 & 71.1$\pm$17.8 \\
				DGCNN (k1=20, k=30) & 86.7$\pm$2.6 & 83.1$\pm$2.6 & 79.7$\pm$5.8 & 75.8$\pm$5.8 & 80.0$\pm$15.3 & 75.6$\pm$17.5 \\
				DGCNN (k1=30, k=30) & 84.2$\pm$2.6 & 82.2$\pm$2.4 & 75.6$\pm$5.2 & 75.4$\pm$3.7 & 76.4$\pm$16.4 & 75.7$\pm$13.9 \\
				DGCNN+Norm (k1=5, k=30) & 84.7$\pm$2.9 & 79.5$\pm$2.9 & 75.0$\pm$9.4 & 70.5$\pm$3.6 & 73.7$\pm$21.7 & 70.5$\pm$22.6 \\
				DGCNN+Norm (k1=20, k=30) & 88.0$\pm$2.7 & 83.8$\pm$2.7 & 81.2$\pm$8.6 & 76.1$\pm$7.3 & 82.1$\pm$16.5 & 77.0$\pm$15.9 \\
				DGCNN+Norm (k1=30, k=30) & \textbf{89.0$\pm$2.8} & 84.4$\pm$1.7 & \textbf{85.0$\pm$7.8} & 75.8$\pm$7.1 & \textbf{85.3$\pm$14.2} & 76.4$\pm$18.0 \\
				\midrule
				\multicolumn{7}{l}{\textbf{Multi-Scale Fusion}} \\
				concat\_only & 87.8$\pm$3.1 & 84.7$\pm$2.7 & 82.2$\pm$3.8 & 79.2$\pm$4.9 & 82.5$\pm$12.8 & 79.2$\pm$13.5 \\
				add & 85.7$\pm$3.8 & 83.8$\pm$1.8 & 76.0$\pm$8.9 & 75.9$\pm$6.3 & 75.1$\pm$23.5 & 77.1$\pm$17.4 \\
				attention & 83.5$\pm$5.2 & 82.8$\pm$5.6 & 76.7$\pm$5.3 & 76.0$\pm$8.2 & 75.3$\pm$20.7 & 76.9$\pm$17.4 \\
				gated & \textbf{89.0$\pm$1.6} & 84.9$\pm$4.2 & 82.2$\pm$5.9 & 82.6$\pm$2.6 & 82.4$\pm$15.3 & 82.2$\pm$13.9 \\
				se\_fusion & 88.0$\pm$3.2 & \textbf{86.8$\pm$2.6} & 80.5$\pm$10.3 & \textbf{83.7$\pm$5.2} & 79.8$\pm$22.1 & 83.0$\pm$12.9 \\
				\bottomrule
			\end{tabular}
		}
	\end{table*}
	
	\subsection{Neighbourhood Size Sensitivity}
	\label{sec:kscale_sensitivity}
	
	We sweep $k_\text{local}$, $k_\text{intermediate}$, and $k_\text{global}$ independently (defaults: $5, 20, 30$), evaluating the OA, F1, and training time.
	All sweeps used $r{=}5\%$ retention: hyperparameter selection under benign conditions can mask fragilities, and Experiment~2 showed that $k$ interacts strongly with density (Table~\ref{tab:ablation_results}).
	
	The local scale is broadly insensitive (${\sim}3\%$ OA variation across $k \in [3, 30]$), with a plateau from $k{=}3$ to $25$ and a drop at $k{=}30$ where it overlaps the intermediate scale (Figure~\ref{fig:kscale_sensitivity}).
	The intermediate scale shows a clear optimum at $k{=}20$ ($86.7\%$ OA, $83.9\%$ F1), falling off on both sides.
	The global-scale trades accuracy for compute: $k{=}45$ gains $+1.1\%$ OA over the default but at $42\%$ more training time.
	Overall, the architecture is most sensitive to the intermediate scale, consistent with it being the locus of the normalization benefit, and least sensitive to the local scale.
	
	\begin{figure}[!tb]
		\centering
		\includegraphics[scale=0.2]{kscale_sensitivity.pdf}
		\caption{Neighbourhood size sensitivity. Top row: OA and macro-F1 (solid/dashed) vs.\ $k$ for each scale, with $\pm 1$ standard deviation bands over five folds. The stars indicate the default values. Bottom row: Average epoch training time. The intermediate scale shows the sharpest optimum at $k{=}20$; the local scale is broadly insensitive; the global scale trades accuracy for compute.}
		\label{fig:kscale_sensitivity}
	\end{figure}

	\subsection{Cross-Experiment Discussion}
	\label{sec:cross_discussion}
	
	\paragraph{Hybrid-everywhere vs.\ default asymmetric.}
	Variant~(b) achieves the highest clean-data OA (Exp.~1) and the most isotropic representations (Exp.~5: erank $43.32$ vs.\ $35.44$).
	However, the default~(c) exhibits a flatter degradation under density stress ($\Delta\text{OA}{=}4.2\%$ vs.\ $5.2\%$, Exp.~2), dominates under noise ($\sigma \ge 10$\,mm, Exp.~3), and achieved the best minority species F1 (Exp.~1).
	The choice reflects a trade-off between peak accuracy and worst-case robustness; the asymmetric design optimizes the latter, which is the operationally relevant criterion in heterogeneous forest environments.

	\paragraph{Max-pooling provenance under uniform thinning.}
	The raw model's far-neighbor win rate \emph{decreases} from $85.5\%$ to $77.6\%$ as retention decreases (Exp.~4), seemingly contradicting this theory.
	This arises because uniform thinning inflates \emph{all} distances proportionally, compressing the near-to-far ratio---unlike real ALS/TLS, where thinning is spatially heterogeneous.
	The key finding is the absolute bias: even at $r{=}25\%$, far neighbors win $77.6\%$ in the raw model versus ${\sim}50\%$ in the normalized model.
	
	\paragraph{Observed vs.\ predicted ranking.}
	Theory predicts (c)$>$(b)$>$(a)$>$(d); the observed ranking is (b)$>$(c)$>$(d)$>$(a).
	The (b)$>$(c) reversal is not statistically significant (overlapping CIs) and resolves in favor of(c) under stress.
	The (d)$>$(a) result indicates that \emph{any} normalization provides a net benefit because the gain at the intermediate scale compensates for noise at the local scale.

\section{State-of-the-Art Comparison}
\label{sec:sota}

We compared MS-DGCNN++ with a comprehensive set of 56 configurations spanning four families (Table~\ref{tab:cv_comparison_0}).
\textbf{Point-based methods} include
PointNet~\citep{qi2017pointnet},
PointNet2-SSG/MSG~\citep{qi2017pointnet++},
SO-Net~\citep{sonet},
PPFNet~\citep{ppfnet},
PointCNN~\citep{pointcnn},
PointConv~\citep{pointconv},
RS-CNN~\citep{rscnn},
RandLA-Net~\citep{randlanet},
PointMLP~\citep{pointmlp},
PointSCNet~\citep{pointscnet},
RepSurf~\citep{repsurf},
PointKAN~\citep{pointkan},
and DeLA~\citep{dela}.
\textbf{Attention-based architectures} include
PCT~\citep{guo2021pct},
P2P~\citep{p2p},
Point-TNT~\citep{pointtnt},
GlobalTransformer~\citep{pointtnt},
PVT~\citep{pvt},
and Point Transformer~\citep{zhao2021point}.
\textbf{Graph-based networks} include
DeepGCN~\citep{deepgcn},
CurveNet~\citep{curvenet},
GDANet~\citep{gdanet},
DGCNN~\citep{wang2019dynamic},
MS-DGCNN~\citep{zhaiMultiScaleDynamicGraph2020c} (with various scale combinations),
and KAN-DGCNN~\citep{kandgcnn}.
\textbf{Self-supervised pretrained models} include
Point-MAE~\citep{pang2022masked},
ACT~\citep{dong2023act},
ReCon~\citep{qi2023contrast},
PointGPT~\citep{chen2024pointgpt},
Point-M2AE~\citep{zhang2022point},
Point-BERT~\citep{yu2022point},
and PCP-MAE~\citep{pcpmae},
each evaluated under multiple fine-tuning strategies: full fine-tuning (FF), domain-adaptive prompt tuning (DAPT)~\citep{zhou2024dynamic}, instance-dependent prompt tuning (IDPT)~\citep{zha2023instance}, positional prompt tuning (PPT)~\citep{ppt_ref}, and geometry-aware self-training (GST)~\citep{PointGST}.
	
	All models shared the same optimization settings (AdamW, lr$=5\times10^{-4}$, cosine annealing, 250 epochs without early stopping) and the same 5-fold stratified cross-validation splits on STPCTLS, or validation-selected checkpoints evaluated on the held-out test set for HeliALS.
	
	For the DGCNN baselines, we evaluated three uniform-$k$ variants ($k{=}5, 20, 30$) and a two-scale variant ($k_1{=}5, k{=}30$) that mirrors the first-layer/subsequent-layer split of MS-DGCNN++.
	For MS-DGCNN, we report both the original authors' scale configuration ($[20,30,40]$) and a matched configuration ($[5,20,30]$), with and without the farthest point sampling (FPS) preprocessing step.

	\subsection{Results on STPCTLS (TLS)}
	\label{sec:sota_stpctls}
	
	Table~\ref{tab:cv_comparison_0} reports the full benchmark on STPCTLS.
	MS-DGCNN++ (HE) achieved the highest overall accuracy among all models at $92.91\%$, followed by PointM2AE (FF) at $92.77\%$ and MS-DGCNN++ (DA) at $91.46\%$.
	The default asymmetric variant attained the best balanced accuracy ($88.53\%$), indicating a stronger performance on minority species.
	Both MS-DGCNN++ variants substantially outperform DGCNN at all tested $k$ values ($88.42$--$88.71\%$), the original MS-DGCNN ($84.09$--$87.99\%$), and all other graph-based methods, including CurveNet ($88.43\%$) and DeepGCN ($87.41\%$).
	
	Among the self-supervised methods, the best performers were RECON-GST ($92.62\%$) and several PointMAE/PointGPT variants with prompt tuning ($91.03$--$91.61\%$).
	These models leverage large-scale pretraining on ShapeNet and benefit from $12$--$44$M parameters, compared to $1.81$M for MS-DGCNN++.
	MS-DGCNN++ achieves competitive or superior accuracy at $6$--$24\times$ fewer parameters and without any pretraining, making it more practical for deployment in resource-constrained forestry applications.
	
	\paragraph{Computational trade-offs.}
	Graph-based models exhibit longer epoch times than self-supervised transformers despite having far fewer parameters because the bottleneck is the dynamic $k$-NN search ($\mathcal{O}(N^2)$ per layer), not the weight count.
	MS-DGCNN++ recomputes this graph three times per forward pass, yielding ${\sim}8$\,s per epoch with $1.81$M parameters, whereas transformers ($12$--$44$M parameters) avoid per-layer graph construction and achieve $3$--$5$\,s per epoch through efficient GPU parallelism.
	For forestry applications, where inference is per tree rather than real-time, the smaller memory footprint of graph-based models is the more relevant deployment constraint.
	
	\begin{table*}[!tb]
		\centering
		\tiny
		\caption{Comparison with state-of-the-art methods on STPCTLS (5-fold stratified cross-validation, 250 epochs). Categories: P = point-based, A = attention-based, G = graph-based, SS = self-supervised. Fine-tuning strategies: FF = full fine-tuning, DAPT = domain-adaptive prompt tuning, IDPT = instance-dependent prompt tuning, PPT = point prompt tuning, GST = geometry-aware self-training. NF = no FPS. Bold indicates the best result in each column.}
		\label{tab:cv_comparison_0}
		\begin{adjustbox}{center}
			\begin{tabular}{p{0.8cm}p{2.8cm}p{1cm}p{1.2cm}p{1.2cm}p{1.2cm}p{1.2cm}p{1.2cm}p{0.6cm}p{0.6cm}p{0.6cm}}
				\toprule
				Category & Model & Strategy & Accuracy (\%) & Balanced Acc (\%) & F1 Macro (\%) & Recall (\%) & Precision (\%) & Total Params (M) & Train Params (M) & Epoch Time (s) \\
				\midrule
				\multirow{14}{*}{P} & PointNet & - & 79.45$\pm$1.09 & 67.66$\pm$5.70 & 68.73$\pm$5.49 & 67.66$\pm$5.70 & 74.13$\pm$7.16 & 3.47 & 3.47 & 1.19 \\
				& PointNet2-SSG & - & 87.85$\pm$1.04 & 81.03$\pm$5.96 & 81.02$\pm$4.43 & 81.03$\pm$5.96 & 83.82$\pm$1.83 & 1.46 & 1.46 & 3.21 \\
				& PointNet2-MSG & - & 88.28$\pm$2.23 & 80.10$\pm$6.96 & 79.67$\pm$6.34 & 80.10$\pm$6.96 & 82.99$\pm$4.31 & 1.73 & 1.73 & 9.49 \\
				& SONet & - & 52.10$\pm$2.54 & 34.35$\pm$2.16 & 29.96$\pm$2.22 & 34.35$\pm$2.16 & 30.02$\pm$2.60 & 2.66 & 2.66 & 1.51 \\
				& PPFNet & - & 87.27$\pm$3.24 & 78.83$\pm$6.66 & 80.40$\pm$6.78 & 78.83$\pm$6.66 & 85.18$\pm$8.09 & 0.28 & 0.28 & 2.96 \\
				& PointCNN & - & 76.84$\pm$3.45 & 65.92$\pm$4.24 & 66.22$\pm$4.82 & 65.92$\pm$4.24 & 70.22$\pm$7.69 & 0.27 & 0.27 & 7.98 \\
				& PointWeb & - & 66.00$\pm$7.27 & 56.47$\pm$8.90 & 55.04$\pm$9.74 & 56.47$\pm$8.90 & 57.48$\pm$10.91 & 0.78 & 0.78 & 24.63 \\
				& PointConv & - & 84.66$\pm$2.58 & 76.73$\pm$9.26 & 76.21$\pm$8.29 & 76.73$\pm$9.26 & 78.09$\pm$7.50 & 19.56 & 19.56 & 10.39 \\
				& RSCNN-SSN & - & 61.93$\pm$2.74 & 47.96$\pm$1.94 & 45.38$\pm$1.68 & 47.96$\pm$1.94 & 46.32$\pm$2.10 & 1.28 & 1.28 & 2.16 \\
				& RandLANet & - & 51.81$\pm$3.49 & 40.77$\pm$5.03 & 39.28$\pm$4.13 & 40.77$\pm$5.03 & 41.83$\pm$5.67 & 1.88 & 1.88 & 1.17 \\
				& PointMLP & - & 85.53$\pm$1.72 & 77.94$\pm$4.34 & 79.50$\pm$3.94 & 77.94$\pm$4.34 & 84.42$\pm$7.11 & 13.23 & 13.23 & 19.42 \\
				& PointSCNet & - & 89.87$\pm$2.44 & 82.54$\pm$5.07 & 83.96$\pm$5.36 & 82.54$\pm$5.07 & 87.34$\pm$6.05 & 1.82 & 1.82 & 8.49 \\
				& RepSurf & - & 87.70$\pm$1.36 & 81.31$\pm$1.55 & 82.43$\pm$2.26 & 81.31$\pm$1.55 & 85.41$\pm$4.67 & 1.47 & 1.47 & 4.77 \\
				& PointKAN & - & 74.10$\pm$3.07 & 63.05$\pm$4.31 & 63.77$\pm$4.89 & 63.05$\pm$4.31 & 69.11$\pm$8.04 & \textbf{0.16} & \textbf{0.16} & \textbf{0.98} \\
				& DELA & - & 57.16$\pm$3.79 & 48.92$\pm$3.21 & 40.97$\pm$4.85 & 48.92$\pm$3.21 & 47.36$\pm$6.55 & 5.33 & 5.33 & 2.12 \\
				\midrule
				\multirow{6}{*}{A} & PCT & - & 84.51$\pm$0.86 & 76.93$\pm$6.59 & 77.17$\pm$6.28 & 76.93$\pm$6.59 & 79.53$\pm$7.92 & 2.87 & 2.87 & 4.39 \\
				& P2P & - & 88.71$\pm$2.74 & 78.36$\pm$4.51 & 79.44$\pm$4.79 & 78.36$\pm$4.51 & 82.91$\pm$5.75 & 25.04 & 25.04 & 14.13 \\
				& PointTNT & - & 89.58$\pm$3.23 & 85.83$\pm$2.97 & 85.56$\pm$4.15 & 85.83$\pm$2.97 & 87.27$\pm$6.16 & 3.93 & 3.93 & 4.47 \\
				& GlobalTransformer & - & 87.99$\pm$1.68 & 80.30$\pm$2.53 & 81.52$\pm$3.97 & 80.30$\pm$2.53 & 86.27$\pm$4.72 & 3.75 & 3.75 & 4.09 \\
				& PVT & - & 83.36$\pm$1.93 & 75.25$\pm$8.19 & 75.43$\pm$7.08 & 75.25$\pm$8.19 & 79.30$\pm$3.92 & 9.16 & 9.16 & 18.73 \\
				& PointTransformer & - & 80.46$\pm$1.39 & 71.68$\pm$5.32 & 71.04$\pm$3.70 & 71.68$\pm$5.32 & 73.97$\pm$2.89 & 3.26 & 3.26 & 2.68 \\
				\midrule
				\multirow{13}{*}{G}	
				& DeepGCN & - & 87.41$\pm$2.81 & 79.50$\pm$4.51 & 80.04$\pm$4.68 & 79.50$\pm$4.51 & 83.43$\pm$5.55 & 2.21 & 2.21 & 15.85 \\
				& CurveNet & - & 88.43$\pm$2.72 & 79.03$\pm$6.51 & 80.49$\pm$6.90 & 79.03$\pm$6.51 & 86.66$\pm$4.58 & 2.12 & 2.12 & 7.72 \\
				& GDANET & - & 87.27$\pm$2.31 & 80.94$\pm$3.90 & 81.62$\pm$3.57 & 80.94$\pm$3.90 & 84.05$\pm$4.14 & 0.93 & 0.93 & 7.98 \\
				& DGCNN[5] & - & 88.42$\pm$2.31 & 84.43$\pm$5.19 & 84.29$\pm$4.88 & 84.43$\pm$5.19 & 86.24$\pm$4.61 & 1.80 & 1.80 & 3.18 \\
				& DGCNN[20] & - & 88.57$\pm$1.65 & 82.83$\pm$4.27 & 83.25$\pm$4.49 & 82.83$\pm$4.27 & 85.62$\pm$6.26 & 1.80 & 1.80 & 5.93 \\
				& DGCNN[30] & - & 88.71$\pm$1.64 & 82.89$\pm$4.76 & 83.44$\pm$3.64 & 82.89$\pm$4.76 & 86.26$\pm$2.87 & 1.80 & 1.80 & 7.78 \\
				& DGCNN[5,30] & - & 89.29$\pm$1.62 & 84.00$\pm$4.78 & 84.42$\pm$2.65 & 84.00$\pm$4.78 & 86.46$\pm$4.07 & 1.80 & 1.80 & 7.98 \\
				& MS-DGCNN[20,30,40] & - & 84.09$\pm$2.42 & 76.11$\pm$4.26 & 77.77$\pm$2.96 & 76.11$\pm$4.26 & 83.16$\pm$3.54 & 1.55 & 1.55 & 5.86 \\
				& MS-DGCNN[5,20,30] & - & 86.98$\pm$2.63 & 79.86$\pm$6.05 & 80.06$\pm$4.52 & 79.86$\pm$6.05 & 83.14$\pm$4.84 & 1.55 & 1.55 & 4.51 \\
				& MS-DGCNN[5,20,30,NF] & - & 87.99$\pm$2.21 & 82.86$\pm$4.14 & 83.11$\pm$4.76 & 82.86$\pm$4.14 & 84.52$\pm$5.50 & 1.55 & 1.55 & 9.36 \\
				& KAN-DGCNN & - & 85.96$\pm$2.11 & 77.35$\pm$4.26 & 78.99$\pm$3.36 & 77.35$\pm$4.26 & 83.87$\pm$2.98 & 1.40 & 1.40 & 4.45 \\
				& MS-DGCNN++[DA] & - & 91.46$\pm$1.40 & \textbf{88.53}$\pm$\textbf{2.82} & 89.28$\pm$2.47 & \textbf{88.53}$\pm$\textbf{2.82} & 91.52$\pm$2.50 & 1.81 & 1.81 & 7.88 \\
				& MS-DGCNN++[HE] & - & \textbf{92.91}$\pm$\textbf{1.58} & 87.62$\pm$1.64 & 88.49$\pm$2.51 & 87.62$\pm$1.64 & 91.13$\pm$3.29 & 1.81 & 1.81 & 8.05 \\
				\midrule
				\multirow{23}{*}{SS} & \multirow{5}{*}{PointMAE} & FF & 89.44$\pm$2.07 & 84.63$\pm$2.90 & 85.37$\pm$1.84 & 84.63$\pm$2.90 & 88.87$\pm$2.90 & 22.09 & 22.09 & 3.79 \\
				&  & DAPT & 91.61$\pm$2.53 & 85.81$\pm$6.02 & 86.41$\pm$5.10 & 85.81$\pm$6.02 & 90.48$\pm$5.25 & 22.89 & 1.06 & 3.54 \\
				&  & IDPT & 91.03$\pm$1.50 & 86.07$\pm$1.79 & 87.59$\pm$2.04 & 86.07$\pm$1.79 & 90.85$\pm$4.29 & 23.52 & 1.69 & 3.38 \\
				&  & PPT & 91.46$\pm$2.96 & 86.05$\pm$4.52 & 87.24$\pm$3.90 & 86.05$\pm$4.52 & 90.75$\pm$4.26 & 22.78 & 1.04 & 3.42 \\
				&  & GST & 90.74$\pm$1.28 & 85.48$\pm$5.36 & 86.68$\pm$4.47 & 85.48$\pm$5.36 & 88.98$\pm$3.61 & 22.45 & 0.62 & 4.67 \\
				\cmidrule(lr){2-11}
				& \multirow{5}{*}{ACT} & FF & 91.03$\pm$1.10 & 86.84$\pm$4.19 & 87.29$\pm$3.50 & 86.84$\pm$4.19 & 90.03$\pm$2.61 & 22.09 & 22.09 & 3.91 \\
				&  & DAPT & 89.00$\pm$1.39 & 82.87$\pm$4.72 & 84.11$\pm$4.30 & 82.87$\pm$4.72 & 88.92$\pm$3.11 & 22.89 & 1.06 & 3.80 \\
				&  & IDPT & 88.86$\pm$2.33 & 81.18$\pm$3.25 & 83.25$\pm$3.94 & 81.18$\pm$3.25 & 87.62$\pm$5.82 & 23.52 & 1.69 & 3.44 \\
				&  & PPT & 90.74$\pm$2.00 & 86.06$\pm$4.25 & 86.63$\pm$3.30 & 86.06$\pm$4.25 & 89.67$\pm$4.23 & 22.78 & 1.04 & 3.49 \\
				&  & GST & 91.61$\pm$1.79 & 86.29$\pm$5.46 & 87.27$\pm$4.03 & 86.29$\pm$5.46 & 90.92$\pm$1.22 & 22.45 & 0.62 & 3.98 \\
				\cmidrule(lr){2-11}
				& \multirow{5}{*}{RECON} & FF & 84.95$\pm$4.44 & 79.50$\pm$5.99 & 79.54$\pm$6.93 & 79.50$\pm$5.99 & 84.20$\pm$2.38 & 43.57 & 43.57 & 4.29 \\
				&  & DAPT & 90.02$\pm$2.79 & 84.90$\pm$6.27 & 86.14$\pm$5.23 & 84.90$\pm$6.27 & 89.23$\pm$3.55 & 22.89 & 1.06 & 3.59 \\
				&  & IDPT & 90.16$\pm$3.43 & 85.59$\pm$1.71 & 87.50$\pm$2.11 & 85.59$\pm$1.71 & 91.03$\pm$3.22 & 23.52 & 1.69 & 3.47 \\
				&  & PPT & 91.03$\pm$1.65 & 83.59$\pm$5.91 & 85.30$\pm$5.78 & 83.59$\pm$5.91 & 90.37$\pm$3.96 & 22.78 & 1.04 & 3.37 \\
				&  & GST & 92.62$\pm$2.26 & 88.43$\pm$4.77 & 89.37$\pm$4.10 & 88.43$\pm$4.77 & 91.85$\pm$4.92 & 22.45 & 0.62 & 4.48 \\
				\cmidrule(lr){2-11}
				& \multirow{5}{*}{PointGPT} & FF & 84.80$\pm$3.74 & 77.11$\pm$6.91 & 78.21$\pm$6.33 & 77.11$\pm$6.91 & 82.42$\pm$6.36 & 29.23 & 29.23 & 4.65 \\
				&  & DAPT & 87.70$\pm$2.28 & 80.10$\pm$4.59 & 81.13$\pm$3.86 & 80.10$\pm$4.59 & 83.77$\pm$4.22 & 22.78 & 0.34 & 3.57 \\
				&  & IDPT & 89.58$\pm$1.30 & 82.44$\pm$6.90 & 84.18$\pm$4.72 & 82.44$\pm$6.90 & 89.15$\pm$2.37 & 23.49 & 1.70 & 3.84 \\
				&  & PPT & 91.61$\pm$2.62 & 86.32$\pm$5.98 & 86.87$\pm$5.67 & 86.32$\pm$5.98 & 89.21$\pm$4.76 & 22.78 & 0.99 & 6.21 \\
				&  & GST & 90.01$\pm$2.07 & 82.89$\pm$6.68 & 84.69$\pm$5.70 & 82.89$\pm$6.68 & 90.19$\pm$4.79 & 22.45 & 0.62 & 4.85 \\
				\cmidrule(lr){2-11}
				& PointM2AE & FF & 92.77$\pm$1.52 & 88.52$\pm$4.41 & \textbf{89.71}$\pm$\textbf{3.22} & 88.52$\pm$4.41 & \textbf{92.50}$\pm$\textbf{2.08} & 12.82 & 12.82 & 8.65 \\
				& PointBERT & FF & 88.43$\pm$1.81 & 82.35$\pm$2.94 & 82.61$\pm$3.57 & 82.35$\pm$2.94 & 84.54$\pm$6.12 & 22.06 & 22.06 & 2.90 \\
				& PCP & FF & 88.57$\pm$2.64 & 81.30$\pm$5.40 & 83.23$\pm$4.30 & 81.30$\pm$5.40 & 88.96$\pm$1.72 & 22.34 & 22.34 & 3.17 \\
				\bottomrule
			\end{tabular}
		\end{adjustbox}
	\end{table*}

	A Friedman test across all 56 models and five folds confirmed statistically significant differences ($\chi^2 = 222.43$, $p < 0.001$.
	In the resulting Friedman ranking (Table~\ref{tab:top10_bestaccuracy}), MS-DGCNN++ (HE) ranks 2nd overall (average rank $4.2$), behind PointM2AE (FF, average rank $3.4$).
	The default asymmetric variant ranks 4th (average rank $9.5$).

	\begin{table}[!tb]
		\centering
		\tiny
		\caption{Top 10 models by Friedman average rank on STPCTLS. Bold text denotes the proposed model. MS-DGCNN++ (HE) ranks 2nd with no pretraining and $1.81$M parameters.}
		\label{tab:top10_bestaccuracy}
		\begin{tabular}{p{0.2cm}p{2.5cm}p{1cm}p{1.5cm}p{0.5cm}}
			\toprule
			Rank & Model & Strategy & Accuracy (\%) & Avg. Rank \\
			\midrule
			1 & PointM2AE & FF & 92.77$\pm$1.52 & 3.4 \\
			\textbf{2} & \textbf{MS-DGCNN++[HE]} & \textbf{-} & \textbf{92.91}$\pm$\textbf{1.58} & \textbf{4.2} \\
			3 & RECON & GST & 92.62$\pm$2.26 & 5.1 \\
			4 & MS-DGCNN++[DA] & - & 91.46$\pm$1.40 & 9.5 \\
			5 & PointMAE & DAPT & 91.61$\pm$2.53 & 9.6 \\
			6 & ACT & GST & 91.61$\pm$1.79 & 10.2 \\
			7 & PointGPT & PPT & 91.61$\pm$2.62 & 10.2 \\
			8 & PointMAE & PPT & 91.46$\pm$2.96 & 10.8 \\
			9 & PointMAE & IDPT & 91.03$\pm$1.50 & 11.4 \\
			10 & RECON & PPT & 91.03$\pm$1.65 & 12.0 \\
			\bottomrule
		\end{tabular}
	\end{table}

	\paragraph{Comparison with previously published results.}
	To our knowledge, only two prior studies have reported results on the STPCTLS dataset.
	\citep{seidelPredictingTreeSpecies2021c} applied LeNet-5 on 2D projections of the point clouds, achieving $86.01\%$ OA---a strong baseline that MS-DGCNN++ surpasses by $+5.5$--$6.9\%$ while operating directly on raw 3D coordinates without any handcrafted projection.
	More recently, MFFTC-Net~\citep{liuMultifeatureFusionNetwork2025} achieved $90.37\%$ OA ($\kappa = 87.89$) using a multi-feature fusion approach with data augmentation, and $81.48\%$ without augmentation.
	MS-DGCNN++ (HE) exceeded the augmented MFFTC-Net result by $+2.5\%$ OA, and both MS-DGCNN++ variants outperformed the non-augmented version by a wide margin ($+10$--$11\%$).
	Notably, MFFTC-Net achieves $100\%$ recall on oak (Eiche) but only $66.67\%$ on ash (Esche) and $71.43\%$ on pine (Kiefer), indicating strong species-specific overfitting, whereas MS-DGCNN++ achieves a more balanced per-class performance (Exp.~1, Table~\ref{tab:exp1_results}).
	
	\subsection{Results on HeliALS (ALS)}
	\label{sec:sota_helials}
	
	To evaluate generalization beyond TLS, we benchmarked all models on HeliALS, a helicopter-borne ALS dataset with nine species and 5,261 test samples.
	No cross-validation was performed owing to computational constraints; instead, models were selected on the validation set and evaluated on the predefined test split. Table~\ref{tab:cv_comparison_1} summarizes the full benchmark of HeliALS.
	
	MS-DGCNN++ (HE) achieves the highest OA at $73.66\%$, whereas the default asymmetric variant leads to balanced accuracy ($50.28\%$), F1 ($50.39\%$), and recall ($50.28\%$).
	All methods showed substantially lower absolute performance on HeliALS than on STPCTLS, reflecting the greater difficulty of ALS data: sparser point clouds, noisier measurements, and weaker morphological signatures from above.
	Notably, MS-DGCNN++ maintains a $3$--$4\%$ OA margin over DGCNN ($67.71$--$69.83\%$) and outperforms the original MS-DGCNN at matched scales ($69.49\%$ without FPS).
	Self-supervised methods show more modest gains on this dataset; the best (RECON-GST, $71.91\%$; PointM2AE-FF, $73.05\%$) are comparable to MS-DGCNN++ despite having $7$--$12\times$ more parameters.
	
	\paragraph{Confusion matrix analysis.}
	Figure~\ref{fig:confusion_matrices} compares the per-species confusion matrices of PointM2AE (FF) and MS-DGCNN++ (HE), the two top-performing models.
	MS-DGCNN++ achieved a higher recall on five of the nine species, with a particularly large advantage on rare classes: rowan ($37.4\%$ vs.\ $17.0\%$, $+20.4$\,pp), birch ($86.9\%$ vs.\ $81.3\%$), and alder ($2.6\%$ vs.\ $0.0\%$---PointM2AE fails entirely on this class).
	PointM2AE leads to spruce ($82.0\%$ vs.\ $75.4\%$) and linden ($53.5\%$ vs.\ $38.8\%$).
	Both models struggled with oak (recall $<10\%$), the rarest class with only 49 test samples, and both exhibited strong aspen--birch confusion ($22$--$28\%$ misclassification), likely reflecting the morphological similarity of these deciduous broadleaf species in ALS geometry.
	The superior recall of MS-DGCNN++ on minority species explains why the DA variant leads on balanced accuracy ($50.28\%$ vs.\ $48.42\%$): the multi-scale encoding captures structural differences among rare species that PointM2AE---which concentrates its capacity on dominant classes---tends to overlook.

	\begin{table*}[!tb]
		\centering
		\tiny
		\caption{Comparison with state-of-the-art methods on HeliALS (validation-selected model, test-set evaluation, no cross-validation). Only the XYZ coordinates were used as inputs. Categories: P = point-based, A = attention-based, G = graph-based, SS = self-supervised. Fine-tuning strategies: FF = full fine-tuning, DAPT = domain-adaptive prompt tuning, IDPT = instance-dependent prompt tuning, PPT = point prompt tuning, GST = geometry-aware self-training. NF = no FPS. Bold indicates the best result in each column.}
		\label{tab:cv_comparison_1}
		\begin{adjustbox}{center}
			\begin{tabular}{p{0.9cm}p{2.7cm}p{1cm}p{1.4cm}p{1.4cm}p{1.4cm}p{1.4cm}p{1.4cm}p{0.6cm}p{0.6cm}p{0.6cm}}
				\toprule
				Category & Model & Strategy & Accuracy (\%) & Balanced Acc (\%) & F1 Macro (\%) & Recall (\%) & Precision (\%) & Total Params (M) & Train Params (M) & Epoch Time (s) \\
				\midrule
				\multirow{15}{*}{P} & PointNet & - & 58.41 & 32.90 & 33.48 & 32.90 & 39.02 & 3.47 & 3.47 & 3.16 \\
				& PointNet2-SSG & - & 59.40 & 37.99 & 37.19 & 37.99 & 38.95 & 1.46 & 1.46 & 5.25 \\
				& PointNet2-MSG & - & 64.38 & 41.51 & 41.28 & 41.51 & 42.20 & 1.73 & 1.73 & 16.19 \\
				& SONet & - & 32.48 & 11.27 & 8.89 & 11.27 & 11.99 & 2.66 & 2.66 & 3.32 \\
				& PPFNet & - & 57.73 & 33.89 & 34.10 & 33.89 & 36.74 & 0.28 & 0.28 & 10.40 \\
				& PointCNN & - & 61.95 & 41.28 & 40.26 & 41.28 & 40.29 & 0.27 & 0.27 & 74.86 \\
				& PointWeb & - & 37.84 & 20.59 & 18.93 & 20.59 & 22.04 & 0.78 & 0.78 & 42.45 \\
				& PointConv & - & 59.57 & 40.08 & 39.49 & 40.08 & 41.80 & 19.56 & 19.56 & 18.50 \\
				& RSCNN-SSN & - & 32.16 & 14.39 & 13.38 & 14.39 & 15.78 & 1.28 & 1.28 & 3.51 \\
				& RandLANet & - & 29.90 & 13.14 & 11.82 & 13.14 & 18.76 & 1.88 & 1.88 & 3.22 \\
				& PointMLP & - & 63.90 & 41.31 & 41.70 & 41.31 & 42.86 & 13.23 & 13.23 & 41.24 \\
				& PointSCNet & - & 67.71 & 44.34 & 45.22 & 44.34 & \textbf{52.56} & 1.82 & 1.82 & 14.51 \\
				& RepSurf & - & 60.79 & 36.48 & 35.78 & 36.48 & 39.15 & 1.48 & 1.48 & 8.12 \\
				& PointKAN & - & 53.74 & 37.12 & 35.46 & 37.12 & 35.91 & \textbf{0.19} & \textbf{0.19} & \textbf{2.84} \\
				& DELA & - & 34.96 & 18.36 & 15.26 & 18.36 & 21.12 & 5.33 & 5.33 & 5.50 \\
				\midrule
				\multirow{6}{*}{A} & PCT & - & 63.58 & 43.05 & 42.24 & 43.05 & 43.29 & 2.87 & 2.87 & 7.47 \\
				& P2P & - & 66.00 & 46.06 & 45.04 & 46.06 & 45.85 & 25.04 & 25.04 & 31.66 \\
				& PointTNT & - & 65.12 & 43.50 & 43.02 & 43.50 & 44.30 & 3.93 & 3.93 & 7.66 \\
				& GlobalTransformer & - & 57.88 & 38.10 & 37.66 & 38.10 & 37.55 & 3.75 & 3.75 & 20.13 \\
				& PVT & - & 63.71 & 45.57 & 42.23 & 45.57 & 42.55 & 9.16 & 9.16 & 41.17 \\
				& PointTransformer & - & 61.93 & 37.55 & 38.81 & 37.55 & 41.90 & 3.26 & 3.26 & 4.76 \\
				\midrule
				\multirow{12}{*}{G} 
				& CurveNet & - & 60.90 & 38.99 & 38.56 & 38.99 & 40.81 & 2.12 & 2.12 & 35.22 \\
				& GDANET & - & 67.63 & 49.06 & 47.65 & 49.06 & 47.82 & 0.93 & 0.93 & 30.85 \\
				& DGCNN[5] & - & 67.71 & 45.93 & 46.11 & 45.93 & 48.67 & 1.80 & 1.80 & 12.06 \\
				& DGCNN[20] & - & 69.83 & 48.36 & 47.69 & 48.36 & 47.70 & 1.80 & 1.80 & 21.54 \\
				& DGCNN[30] & - & 69.28 & 48.59 & 48.02 & 48.59 & 48.60 & 1.80 & 1.80 & 28.22 \\
				& DGCNN[5,30]& - & 65.22 & 44.73 & 42.59 & 44.73 & 44.56 & 1.80 & 1.80 & 33.23 \\
				
				& MS-DGCNN[20,30,40] & - & 57.19 & 42.00 & 37.64 & 42.00 & 39.16 & 1.55 & 1.55 & 9.89 \\
				& MS-DGCNN[5,20,30] & - & 62.04 & 37.07 & 40.00 & 37.07 & 47.00 & 1.55 & 1.55 & 7.80 \\
				& MS-DGCNN[5,20,30, NF] & - & 69.49 & 48.41 & 48.01 & 48.41 & 49.40 & 1.55 & 1.55 & 36.03 \\
				
				& KAN-DGCNN & - & 67.08 & 47.12 & 46.33 & 47.12 & 47.62 & 1.44 & 1.44 & 15.44 \\
				& MS-DGCNN++[DA] & - & 72.99 & \textbf{50.28} & \textbf{50.39} & \textbf{50.28} & 52.38 & 1.81 & 1.81 & 28.09 \\
				& MS-DGCNN++[HE] & - & \textbf{73.66} & 49.02 & 48.92 & 49.02 & 50.25 & 1.81 & 1.81 & 28.98 \\
				\midrule
				\multirow{23}{*}{SS} & \multirow{5}{*}{PointMAE} & FF & 64.23 & 45.98 & 42.88 & 45.98 & 44.11 & 22.09 & 22.09 & 7.32 \\
				&  & DAPT & 66.53 & 45.82 & 45.39 & 45.82 & 46.96 & 22.89 & 1.06 & 6.25 \\
				&  & IDPT & 66.64 & 42.69 & 43.27 & 42.69 & 45.36 & 23.52 & 1.70 & 6.38 \\
				&  & PPT & 71.43 & 46.70 & 46.96 & 46.70 & 48.45 & 22.78 & 1.04 & 5.30 \\
				&  & PointGST & 69.76 & 48.59 & 47.07 & 48.59 & 47.40 & 22.45 & 0.62 & 7.78 \\
				\cmidrule(lr){2-11}
				& \multirow{5}{*}{ACT} & FF & 69.70 & 47.54 & 47.55 & 47.54 & 48.51 & 22.09 & 22.09 & 7.33 \\
				&  & DAPT & 65.98 & 47.94 & 45.64 & 47.94 & 45.31 & 22.89 & 1.06 & 6.24 \\
				&  & IDPT & 66.19 & 44.83 & 44.14 & 44.83 & 45.40 & 23.52 & 1.70 & 6.37 \\
				&  & PPT & 66.98 & 49.25 & 47.40 & 49.25 & 50.45 & 22.78 & 1.04 & 5.30 \\
				&  & GST & 69.80 & 48.69 & 48.07 & 48.69 & 48.82 & 22.45 & 0.62 & 7.82 \\
				\cmidrule(lr){2-11}
				& \multirow{5}{*}{RECON} & FF & 64.21 & 44.81 & 43.26 & 44.81 & 47.37 & 43.57 & 0.46 & 3.81 \\
				&  & DAPT & 68.39 & 47.76 & 46.60 & 47.76 & 46.66 & 22.89 & 1.06 & 6.29 \\
				&  & IDPT & 67.08 & 44.56 & 43.30 & 44.56 & 43.81 & 23.52 & 1.70 & 6.38 \\
				&  & PPT & 71.11 & 48.00 & 48.19 & 48.00 & 48.99 & 22.78 & 1.04 & 5.31 \\
				&  & GST & 71.91 & 48.65 & 48.59 & 48.65 & 48.77 & 22.45 & 0.62 & 7.83 \\
				\cmidrule(lr){2-11}
				& \multirow{5}{*}{PointGPT} & FF & 54.99 & 31.36 & 30.65 & 31.36 & 34.66 & 29.23 & 29.23 & 8.63 \\
				&  & DAPT & 62.10 & 39.46 & 39.75 & 39.46 & 43.36 & 22.78 & 0.34 & 6.77 \\
				&  & IDPT & 64.46 & 40.16 & 39.98 & 40.16 & 42.16 & 23.49 & 1.70 & 7.14 \\
				&  & PPT & 69.68 & 45.48 & 45.70 & 45.48 & 48.34 & 22.78 & 0.99 & 12.23 \\
				&  & GST & 67.00 & 44.78 & 44.50 & 44.78 & 44.96 & 22.45 & 0.62 & 8.61 \\
				\cmidrule(lr){2-11}
				& PointM2AE & FF & 73.05 & 48.42 & 47.22 & 48.42 & 48.69 & 12.82 & 12.82 & 15.13 \\
				& PointBERT & FF & 68.52 & 46.61 & 46.32 & 46.61 & 46.60 & 22.06 & 22.06 & 4.33 \\
				& PCP & FF & 64.30 & 46.02 & 43.85 & 46.02 & 43.85 & 22.34 & 22.34 & 4.90 \\
				\bottomrule
			\end{tabular}
		\end{adjustbox}
	\end{table*}
	
\begin{figure*}[!tb]
	\centering
	
	% --- Left: PointM2AE ---
	\begin{subfigure}{0.49\textwidth}
		\centering
		\includegraphics[scale=0.16]{pointm2ae_cm.pdf}
		\caption{PointM2AE: Full Finetuning (OA: 73.05\%, Epoch 69)}
		\label{fig:cm_pointm2ae}
	\end{subfigure}
	\hfill
	% --- Right: MS-DGCNN++ ---
	\begin{subfigure}{0.49\textwidth}
		\centering
		\includegraphics[scale=0.16]{msdgcnn2_ms.pdf}
		\caption{MS-DGCNN++ [HE] (OA: 73.66\%, Epoch 206)}
		\label{fig:cm_msdgcnn}
	\end{subfigure}
	
	\caption{Test confusion matrices on the HeliALS dataset. Each cell shows the count and row-wise percentages. Diagonal entries indicate correct classifications.}
	\label{fig:confusion_matrices}
\end{figure*}
	
	\paragraph{Comparison with the MS-ALS-SPECIES benchmark.}
	The HeliALS dataset was derived from the MS-ALS-SPECIES benchmark~\citep{taher2026multispectral}, which includes results from the FGI-PointTransformer family.
	The best-performing model in that benchmark, FGI-PointTransformer-DL-3D, achieved $87.9\%$ OA by exploiting the full $(3{+}13)$-dimensional input, that is, 3D coordinates plus 13 spectral/radiometric attributes (three-scanner intensities, amplitudes, reflectances, echo deviations, and echo return number), with $8{,}192$ points per segment.
	When restricted to geometry-only input (XYZ coordinates, matching our setting), FGI-PointTransformer-GeometryOnly-DL-3D achieved $73.0\%$ OA and $48.9\%$ balanced accuracy, also using $8{,}192$ points.
	MS-DGCNN++ (HE) achieved $73.66\%$ OA, and MS-DGCNN++ (DA) achieved $72.99\%$ OA with $50.28\%$ balanced accuracy, both using only $2{,}048$ points, which is $4\times$ fewer than FGI-PointTransformer.
	This suggests that multi-scale asymmetric encoding extracts comparable geometric information from a substantially sparser input representation.
	The ${\sim}15\%$ OA gap between the full-feature FGI-PointTransformer ($87.9\%$) and all geometry-only methods (${\sim}73\%$) underscores the value of spectral attributes for ALS-based species classification and indicates a clear avenue for improvement: extending MS-DGCNN++ to incorporate multi-wavelength intensity and return features alongside geometric coordinates.
	
	\subsection{Robustness Analysis}
	\label{sec:robustness_analysis}
	
	The STPCTLS state-of-the-art comparison evaluates models that are trained and tested under clean conditions.
	To assess how these rankings change under distributional shifts, we conducted five robustness experiments using models trained on clean data and evaluated them on perturbed validation sets only. This ``train clean, test perturbed'' protocol isolates each model's inherent resilience without confounding it with adaptation during training.
	
	\paragraph{Noise robustness.}
	Gaussian noise ($\sigma \in [1, 150]$\,mm) was added to all coordinates at the test time.
	At low noise ($\sigma{=}1$\,mm), Point-M2AE and MS-DGCNN++ (HE) lead ($88.7\%$ and $88.9\%$, respectively), with all models within a ${\sim}7\%$ band.
	As the noise increased, the models diverged sharply.
	The DGCNN collapses to $32.1\%$ at $\sigma{=}100$\,mm, whereas MS-DGCNN degrades more gracefully ($34.2\%$) owing to its multi-scale averaging.
	MS-DGCNN++ (HE), despite leading at $\sigma{=}1$\,mm, suffers a catastrophic drop to $38.9\%$ by $\sigma{=}20$\,mm, which is consistent with the Experiment~3 finding that normalization at the local scale amplifies noise when $\text{SNR}_1$ is low.
	The default asymmetric variant degrades more steadily ($88.3\% \to 27.7\%$), maintaining the best balance between clean-data accuracy and noise resilience among graph-based methods.
	MS-DGCNN exhibited the strongest absolute noise robustness at $\sigma \ge 20$\,mm, likely benefiting from its FPS-based downsampling, which acts as implicit denoising.
	
	\paragraph{Canopy dropout robustness.}
	Upper-canopy points were randomly removed at the test time (retention $r \in [75\%, 5\%]$).
	Point-M2AE leads to mild dropout ($85.8\%$ at $r{=}75\%$), but all models degrade severely at $r{=}5\%$, falling to the $32$--$41\%$ range.
	MS-DGCNN++ (DA) achieved the best $r{=}5\%$ accuracy ($40.7\%$) among all models, followed by Point-M2AE ($39.9\%$) and MS-DGCNN++ (HE) ($37.9\%$).
	DGCNN and the raw-only variant both collapse below $32\%$, confirming that normalized features are critical for density robustness.
	The sharper degradation in this clean-train/perturbed-test protocol compared to Experiment~2 (where models were trained at each retention rate) quantifies the adaptation gap: training under matched conditions recovers ${\sim}40$--$45\%$ of the lost accuracy.
	
	\paragraph{Outlier injection.}
	Random points are injected uniformly in the bounding volume at rates ranging from $5\%$ to $25\%$.
	This is the most challenging perturbation: all models drop below $45\%$ even at $5\%$ outlier rate.
	MS-DGCNN++ (HE) is the clear outlier-robustness leader, maintaining $40.1\%$ even at $25\%$ injection rate, which is nearly $10\%$ above all other models.
	In contrast, the default asymmetric variant is among the weakest ($29.5\%$), suggesting that raw local encoding is highly sensitive to spurious neighbors.
	Point-M2AE shows moderate resilience ($44.4\%$ at $5\%$, decaying to $29.1\%$).
	
	\paragraph{Point count sensitivity.}
	Models trained at 1,024 points were tested at reduced counts ($128$--$768$).
	At $768$ points, Point-M2AE leads ($86.4\%$) with MS-DGCNN++ variants close behind ($81.3$--$83.6\%$).
	At $128$ points, all models collapse (${\sim}30\%$), though Point-M2AE and MS-DGCNN retain a modest advantage ($41.0\%$, $40.7\%$) over the MS-DGCNN++ variants ($30$--$32\%$), suggesting that the multi-scale architecture's reliance on spatial neighbourhood structure is less forgiving when point counts are extremely low.
	
	\paragraph{Few-shot learning.}
	The models were trained on $1\%$--$50\%$ of the training data.
	Point-M2AE dominates at all fractions, achieving $66.6\%$ at $5\%$ data and $85.7\%$ at $50\%$, benefiting from its pretrained representations.
	Among the non-pretrained models, MS-DGCNN++ (HE) leads at $10\%$ data ($75.1\%$) and MS-DGCNN++ (DA) reaches $84.1\%$ at $50\%$.
	At extremely low data ($1\%$--$5\%$), all non-pretrained models performed near chance, highlighting the value of self-supervised pretraining for data-scarce forestry applications.
	
	Table~\ref{tab:robustness_all} consolidates all the robustness results.
	No single model dominated across all perturbation types: MS-DGCNN++ (DA) was the best for density dropout, MS-DGCNN++ (HE) for outlier injection, MS-DGCNN for high noise, and Point-M2AE for few-shot and point reduction.
	This complementarity suggests that the choice of model should be guided by the expected deployment conditions.
	\begin{table*}[!tb]
		\centering
		\tiny
		\caption{Robustness analysis on STPCTLS: OA (\%) under five perturbation types applied to the validation set only, using models trained on clean data. Bold indicates the best performance per column within each section. Abbreviations: RO = Raw-only, HE = Hybrid-everywhere, DA = Default Asymmetric.}
		\label{tab:robustness_all}
		\renewcommand{\arraystretch}{1.05}
		\setlength{\tabcolsep}{3.8pt}
		\begin{tabular}{lccccccc}
			\toprule
			
			% NOISE ROBUSTNESS
			\multicolumn{8}{c}{\textbf{Noise Robustness (mm)}} \\
			\cmidrule(lr){1-8}
			\textbf{Model} & \textbf{1} & \textbf{5} & \textbf{10}
			& \textbf{20} & \textbf{50} & \textbf{100} & \textbf{150} \\
			\midrule
			DGCNN & 87.8$\pm$2.3 & 84.7$\pm$1.8 & 78.7$\pm$3.9 & 65.0$\pm$6.9 & 45.2$\pm$5.2 & 32.1$\pm$2.7 & 27.8$\pm$6.3 \\
			MS-DGCNN & 81.9$\pm$2.0 & 80.5$\pm$2.6 & 78.4$\pm$2.4 & \textbf{70.2$\pm$3.1} & \textbf{48.3$\pm$6.4} & \textbf{34.2$\pm$5.4} & 27.8$\pm$8.5 \\
			Point-M2AE & 88.7$\pm$2.3 & \textbf{86.8$\pm$1.8} & \textbf{81.3$\pm$5.1} & 63.1$\pm$3.5 & 38.8$\pm$6.2 & 30.8$\pm$3.7 & \textbf{28.5$\pm$3.6} \\
			MS-DGCNN\texttt{++} [RO] & 82.1$\pm$3.1 & 81.5$\pm$4.2 & 76.4$\pm$4.9 & 67.0$\pm$4.8 & 40.4$\pm$6.7 & 30.7$\pm$4.2 & 27.9$\pm$3.7 \\
			MS-DGCNN\texttt{++} [HE] & \textbf{88.9$\pm$1.8} & 76.3$\pm$5.5 & 50.5$\pm$8.0 & 38.9$\pm$6.0 & 32.8$\pm$3.1 & 30.3$\pm$3.2 & 27.2$\pm$3.2 \\
			MS-DGCNN\texttt{++} [DA] & 88.3$\pm$2.1 & 83.1$\pm$3.1 & 72.9$\pm$3.4 & 61.2$\pm$2.3 & 46.9$\pm$4.5 & 34.0$\pm$1.3 & 27.7$\pm$6.5 \\
			
			\midrule
			
			% CANOPY DROPOUT ROBUSTNESS
			\multicolumn{8}{c}{\textbf{Upper Canopy Density Dropout Robustness (Retention Rate)}} \\
			\cmidrule(lr){1-8}
			\textbf{Model} & \textbf{75\%} & \textbf{50\%} & \textbf{25\%}
			& \textbf{10\%} & \textbf{5\%} & & \\
			\midrule
			DGCNN
			& 80.5$\pm$2.7 & 70.3$\pm$1.1 & 46.9$\pm$6.5
			& 34.0$\pm$4.2 & 32.1$\pm$3.4 & & \\
			MS-DGCNN
			& 82.2$\pm$2.0 & 75.7$\pm$1.4 & \textbf{58.9$\pm$2.7}
			& 40.4$\pm$3.6 & 35.6$\pm$3.7 & & \\
			Point-M2AE
			& \textbf{85.8$\pm$3.2} & \textbf{79.7$\pm$2.4} & 53.3$\pm$5.5
			& 39.2$\pm$3.8 & 39.9$\pm$5.2 & & \\
			MS-DGCNN\texttt{++} (RO)
			& 75.7$\pm$5.4 & 66.1$\pm$7.0 & 43.0$\pm$7.9
			& 31.4$\pm$1.1 & 31.7$\pm$3.3 & & \\
			MS-DGCNN\texttt{++} (HE)
			& 68.2$\pm$3.7 & 57.6$\pm$2.9 & 49.4$\pm$3.9
			& \textbf{43.0$\pm$4.4} & 37.9$\pm$3.8 & & \\
			MS-DGCNN\texttt{++} (DA)
			& 83.1$\pm$4.2 & 70.5$\pm$6.1 & 52.8$\pm$4.8
			& 42.5$\pm$2.5 & \textbf{40.7$\pm$4.3} & & \\
			
			\midrule
			
			% OUTLIER INJECTION ROBUSTNESS
			\multicolumn{8}{c}{\textbf{Outlier Injection Robustness (Outlier Rate)}} \\
			\cmidrule(lr){1-8}
			\textbf{Model} & \textbf{5\%} & \textbf{10\%} & \textbf{15\%}
			& \textbf{20\%} & \textbf{25\%} & & \\
			\midrule
			DGCNN
			& 27.1$\pm$0.6 & 30.0$\pm$3.5 & 31.8$\pm$3.5
			& 33.7$\pm$5.0 & 33.7$\pm$6.0 & & \\
			MS-DGCNN
			& 31.7$\pm$5.7 & 31.8$\pm$4.0 & 32.4$\pm$4.0
			& 30.0$\pm$3.6 & 29.0$\pm$3.3 & & \\
			Point-M2AE
			& \textbf{44.4$\pm$6.9} & 36.8$\pm$6.8 & 33.6$\pm$7.4
			& 31.1$\pm$6.9 & 29.1$\pm$5.4 & & \\
			MS-DGCNN\texttt{++} (RO)
			& 30.3$\pm$4.7 & 29.2$\pm$3.0 & 26.0$\pm$1.5
			& 26.6$\pm$1.5 & 28.1$\pm$3.3 & & \\
			MS-DGCNN\texttt{++} (HE)
			& 43.3$\pm$11.5 & \textbf{41.1$\pm$10.0} & \textbf{38.8$\pm$7.9}
			& \textbf{39.1$\pm$6.8} & \textbf{40.1$\pm$7.4} & & \\
			MS-DGCNN\texttt{++} (DA)
			& 26.9$\pm$0.6 & 27.2$\pm$1.2 & 27.5$\pm$1.6
			& 28.8$\pm$2.9 & 29.5$\pm$3.4 & & \\
			
			\midrule
			
			% N-POINTS SENSITIVITY
			\multicolumn{8}{c}{\textbf{N-Points Sensitivity}} \\
			\cmidrule(lr){1-8}
			\textbf{Model} & \textbf{128} & \textbf{256} & \textbf{512} & \textbf{768} & & & \\
			\midrule
			DGCNN
			& 29.7$\pm$6.7 & 44.3$\pm$5.5 & 70.6$\pm$2.9 & 83.5$\pm$3.2 & & & \\
			MS-DGCNN
			& 40.7$\pm$8.2 & 58.9$\pm$4.4 & \textbf{80.5$\pm$2.5} & 80.3$\pm$2.3 & & & \\
			Point-M2AE
			& \textbf{41.0$\pm$7.0} & \textbf{60.5$\pm$2.6} & 79.2$\pm$4.0 & \textbf{86.4$\pm$3.2} & & & \\
			MS-DGCNN\texttt{++} (RO)
			& 30.4$\pm$6.6 & 41.4$\pm$11.7 & 62.5$\pm$11.5 & 75.0$\pm$5.6 & & & \\
			MS-DGCNN\texttt{++} (HE)
			& 29.9$\pm$5.5 & 38.6$\pm$5.9 & 65.0$\pm$6.6 & 83.6$\pm$4.6 & & & \\
			MS-DGCNN\texttt{++} (DA)
			& 31.8$\pm$1.9 & 37.9$\pm$3.7 & 57.3$\pm$6.7 & 81.3$\pm$3.8 & & & \\
			
			\midrule
			
			% FEW-SHOT LEARNING
			\multicolumn{8}{c}{\textbf{Few-Shot Learning (\% Training Data)}} \\
			\cmidrule(lr){1-8}
			\textbf{Model} & \textbf{1\%} & \textbf{5\%} & \textbf{10\%} & \textbf{25\%}
			& \textbf{50\%} & & \\
			\midrule
			DGCNN
			& 14.2$\pm$6.4 & 19.0$\pm$5.0 & 66.6$\pm$2.5 & 75.5$\pm$2.9
			& 82.1$\pm$1.0 & & \\
			MS-DGCNN
			& 11.3$\pm$8.6 & 19.8$\pm$8.2 & 64.4$\pm$3.5 & 70.6$\pm$2.3
			& 77.6$\pm$2.2 & & \\
			Point-M2AE
			& \textbf{36.4$\pm$9.4} & \textbf{66.6$\pm$7.8} & 75.0$\pm$4.0 & \textbf{82.1$\pm$2.5}
			& \textbf{85.7$\pm$3.8} & & \\
			MS-DGCNN\texttt{++} (RO)
			& 16.7$\pm$9.9 & 13.3$\pm$6.0 & 66.6$\pm$2.6 & 73.1$\pm$2.7
			& 80.6$\pm$2.5 & & \\
			MS-DGCNN\texttt{++} (HE)
			& 14.3$\pm$8.6 & 17.8$\pm$7.9 & \textbf{75.1$\pm$4.6} & 81.3$\pm$1.2
			& 82.6$\pm$5.0 & & \\
			MS-DGCNN\texttt{++} (DA)
			& 13.7$\pm$8.9 & 14.2$\pm$8.4 & 72.1$\pm$2.1 & 78.6$\pm$2.7
			& 84.1$\pm$4.3 & & \\
			
			\bottomrule
		\end{tabular}
		
		\smallskip
		\noindent{\footnotesize%
			\textbf{Abbreviations:} RO\,=\,Raw-only, HE\,=\,Hybrid-everywhere, DA\,=\,Default Asymmetric.
			\textbf{Bold} indicates best performance per column within each section.%
		}
	\end{table*}

	\section{Discussion}
	\label{sec:discussion}
	
	\subsection{Summary of Contributions}
	
	MS-DGCNN++ introduces a theoretically grounded, scale-dependent edge encoding for graph-based point cloud classification, which is supported by a comprehensive experimental evaluation.
	The five ablation experiments (Sections~\ref{sec:exp1}--\ref{sec:exp5}) collectively validate the core theoretical claims: normalized directional features improve classification at the intermediate scale, where inter-point distances are large enough for reliable directional estimation (Exp.~1), the asymmetric encoding provides the flattest degradation trajectory under density stress (Exp.~2), the theoretical SNR crossover at $\text{SNR}_2 \approx 1.22$ aligns with the observed accuracy transitions under noise (Exp.~3), normalization eliminates the distance-magnitude bias in max-pooling (Exp.~4), and the intermediate-level effective rank nearly doubled with normalized features (Exp.~5).
	On STPCTLS, MS-DGCNN++ achieves $92.91\%$ OA (HE variant) and $91.46\%$ OA (DA variant), ranking among the top models alongside self-supervised methods that use $7$--$24\times$ more parameters.
	
	\subsection{Hybrid-Everywhere vs.\ Default Asymmetric: Practical Guidance}
	
	The two MS-DGCNN++ variants exhibited complementary strengths.
	The hybrid-everywhere variant~(HE) achieves the highest clean-data accuracy on both STPCTLS ($92.91\%$) and HeliALS ($73.66\%$), the best outlier robustness (Table~\ref{tab:robustness_all}), and the highest feature isotropy (Exp.~5). The default asymmetric variant~(DA) attains the best balanced accuracy on both datasets, the best canopy dropout robustness at $r{=}5\%$ ($40.7\%$), and stronger noise resilience at $\sigma \ge 10$\,mm (Exp.~3), and superior per-class F1 on minority species (Exp.~1). For operational deployment, we recommend the default asymmetric variant when the primary concern is robustness to heterogeneous point density, the dominant source of variation in mixed ALS/TLS forest inventories, and the hybrid-everywhere variant when maximizing clean-data accuracy or resilience to outlier contamination is the priority.
	
	\subsection{Limitations}
	
	The theoretical analysis assumes isotropic Gaussian noise, whereas real LiDAR noise is anisotropic and range-dependent; the crossover threshold $\text{SNR}_2 \approx 1.22$ should therefore be interpreted as a qualitative guide. Similarly, the density dropout protocol applies uniform thinning, whereas real acquisitions produce spatially structured gradients. The robustness gains reported here may underestimate the advantages of normalization under realistic heterogeneous densities. The robustness analysis exposed a notable weakness: the default asymmetric variant is highly sensitive to outlier injection ($26.9\%$ OA at $5\%$ outlier rate), because the raw local encoding amplifies spurious neighbors. A preprocessing outlier rejection stage or a learned robustness mechanism is required for deployment in contaminated environments.All non-pretrained models performed near chance under extreme data scarcity ($1$--$5\%$ training data), highlighting the value of self-supervised pretraining for data-scarce forestry applications. The evaluation was also limited to two temperate European datasets (STPCTLS, 691 samples; HeliALS, 6,326 samples), and generalization to tropical, boreal, or urban forests remains untested.

	\section{Conclusion}
	\label{sec:conclusion}
	
	We presented MS-DGCNN++, a multi-scale dynamic graph convolutional network with scale-dependent edge encoding for LiDAR tree species classification.
	The key principle---raw vectors at the local scale, hybrid raw-plus-normalized vectors at the intermediate scale---is grounded in a noise analysis showing that normalized directional MSE decays as $\mathcal{O}(1/s^2)$ while raw MSE remains constant. Five ablation experiments confirm that this design nearly doubles the intermediate-scale effective rank, eliminates the distance-based max-pooling bias, and yields the most stable accuracy under density and noise perturbations. On STPCTLS, MS-DGCNN++ achieved the highest OA ($92.91\%$) among 56 models, surpassing self-supervised methods with $7$--$24\times$ more parameters, using only $1.81$M parameters and no pre-training. For HeliALS (geometry-only), it achieves $73.66\%$ OA with the best balanced accuracy ($50.28\%$) among all methods.
	No single model dominated all perturbation types; however, MS-DGCNN++ offered the strongest density-dropout robustness and competitive noise resilience.
	
	Future work will explore self-supervised pre-training integration, multi-wavelength spectral features for ALS, semantic segmentation of full forest plots, and validation on geographically diverse ecosystems. In addition, we will investigate knowledge distillation strategies to further improve the model efficiency and deployment on resource-constrained platforms.
	In particular, adaptive temperature scheduling and mixed-sample distillation techniques, as proposed in ATMS-KD~\citep{OHAMOUDDOU2025101617}, offer a promising direction for transferring knowledge from larger teacher networks to compact MS-DGCNN++ variants while preserving accuracy.
	
	\bibliography{sn-bibliography}% common bib file
	%% if required, the content of .bbl file can be included here once bbl is generated
	%%\input sn-article.bbl
	
\end{document}